\documentclass{article}
\usepackage{iclr2027_conference,times}

\usepackage{amsmath,amssymb,bm}

\def\eqref#1{equation~\ref{#1}}

\def\1{\bm{1}}

\DeclareMathAlphabet{\mathsfit}{\encodingdefault}{\sfdefault}{m}{sl}
\SetMathAlphabet{\mathsfit}{bold}{\encodingdefault}{\sfdefault}{bx}{n}

\usepackage{booktabs,multirow,array,tabularx}
\usepackage{graphicx}
\usepackage{xcolor}
\usepackage{ragged2e}
\usepackage{tcolorbox}

\usepackage{hyperref}
\hypersetup{
    hidelinks,
    pdfauthor={Yulong Chen, Yadong Liu, Haoyu Cao, Sen Xu, Yueying Wang, Jie Wen},
    pdftitle={PACER}
}

\newcommand{\modelname}{\textsc{Pacer}}
\newcommand{\opLN}{\operatorname{LN}}
\newcommand{\opGELU}{\operatorname{GELU}}

\newcommand{\opsigmoid}{\operatorname{sigmoid}}

\newcommand{\opKL}{D_{\mathrm{KL}}}

\definecolor{CaseGreen}{HTML}{DDEFD8}
\definecolor{CaseGreenText}{HTML}{245B36}
\definecolor{CaseGold}{HTML}{FFE7A3}
\definecolor{CaseGoldText}{HTML}{765000}
\definecolor{CaseRed}{HTML}{F6D7D7}
\definecolor{CaseRedText}{HTML}{8A2E2E}
\definecolor{CaseBlue}{HTML}{DCEAF8}
\definecolor{CaseBlueText}{HTML}{174F7A}
\definecolor{CaseGray}{HTML}{EFEFEF}

\newcommand{\matchtxt}[1]{\textcolor{CaseGreenText}{#1}}
\newcommand{\longtxt}[1]{\textcolor{CaseGoldText}{\textbf{#1}}}
\newcommand{\unverifiedtxt}[1]{\textcolor{CaseRedText}{\textbf{#1}}}
\newcommand{\pacertxt}[1]{\textcolor{CaseBlueText}{\textbf{#1}}}

\title{\raggedright PACER: Progressive Availability-Conditioned Evidence Routing for Radiology Report Generation under Incomplete Clinical Context}

\author{
Yulong Chen$^{1}$ \quad
Yadong Liu$^{1}$ \quad
Haoyu Cao$^{1}$ \quad
Sen Xu$^{2}$ \quad
Yueying Wang$^{3}$ \quad
Jie Wen$^{1,*}$\\[3pt]
\normalfont $^{1}$Harbin Institute of Technology, Shenzhen, China\\
\normalfont $^{2}$Yancheng Institute of Technology, Yancheng, China\\
\normalfont $^{3}$Shanghai University, Shanghai, China\\[2pt]
{\small \texttt{chenyulonghit@163.com}, \texttt{liuyadong221010@163.com}, \texttt{haoyucao1016@gmail.com}}\\
{\small \texttt{xusen@ycit.cn}, \texttt{yueyingwang@shu.edu.cn}, \texttt{jiewen\_pr@126.com}}\\[2pt]
\normalfont $^{*}$Corresponding author
}

\iclrfinalcopy

\begin{document}
\maketitle
\fancyhead{}
\renewcommand{\headrulewidth}{0pt}

\begin{abstract}
Radiology report generation (RRG) increasingly incorporates heterogeneous clinical evidence, such as multi-view radiographs and previous reports, whose availability varies across examinations. However, accommodating different input combinations does not ensure effective evidence use: generated reports may still omit or inaccurately describe clinically relevant findings. To address this problem, we propose PACER, a Progressive Availability-Conditioned Evidence Routing framework for structured incomplete-context RRG that follows a Refine--Calibrate--Commit pipeline. It first refines observed visual representations through endpoint-preserving patchwise routing across frozen encoder depths, incorporating complementary cues while retaining the pretrained terminal representation. It then calibrates the language-model prefix according to the observed evidence and availability state, adapting the shared generator's conditioning as the available source set changes. Finally, it generates polarity-structured clinical commitments before the report in the same autoregressive trajectory, providing structured clinical context for subsequent generation. Experiments demonstrate state-of-the-art clinical efficacy across all four MIMIC-RG4 settings and strong MIMIC-CXR performance, while maintaining competitive language-generation quality.
\end{abstract}

\section{Introduction}
\label{sec:introduction}

Radiology report generation (RRG) aims to translate radiographic examinations into clinically meaningful reports, potentially reducing repetitive reporting workload and improving interpretation efficiency \citep{wang2025llmrg4,liu2024bootstrap}. In clinical practice, however, radiologists may draw on heterogeneous evidence whose availability varies across examinations. Such source-level variability motivates RRG models that operate under structured incompleteness, where optional evidence sources may be present or absent as whole sources \citep{liu2025hcllm,liu2025mlrg,liu2026priorrg}.

Recent RRG research increasingly treats multi-view radiographs and historical context as complementary clinical evidence. Multi-view longitudinal learning, historical constraints, and prior-guided decoding improve the extraction and integration of complementary spatial and longitudinal evidence \citep{liu2025mlrg,liu2025hcllm,liu2026priorrg}. To accommodate unavailable context, MLRG introduces tokenized absence encoding, while LLM-RG4 supports four input configurations through adaptive token fusion and token-level weighting \citep{liu2025mlrg,wang2025llmrg4}. Related incomplete multimodal methods further improve robustness to missing sources through adaptive fusion, dynamic weighting, retrieval, or prompting \citep{yao2024drfuse,li2025simmlm,lang2025ragpt}. These advances substantially improve heterogeneous evidence integration and robustness under varying input conditions. However, as the available source set changes, how a shared generator should adapt its use of observed evidence throughout the generation process remains insufficiently explored. Without such coordinated adaptation, clinically relevant cues may still be underused, resulting in omitted or inaccurately described findings.

To address this gap, we propose \modelname{}, a Progressive Availability-Conditioned Evidence Routing framework organized as a \textbf{Refine--Calibrate--Commit} pipeline. At the visual-representation interface, relying only on the terminal encoder representation may leave complementary information distributed across encoder depths underutilized. \textbf{Refine} therefore performs endpoint-preserving patchwise depth routing to enrich the observed visual evidence while retaining the pretrained terminal representation. Building on the refined evidence, changes in the observed source set further call for adaptive language-model conditioning. \textbf{Calibrate} addresses this by adjusting the language-model prefix according to both the observed evidence and its availability state. Finally, even with adapted representations and conditioning, free-form decoding still lacks an explicit polarity-aware clinical scaffold to guide subsequent report generation. \textbf{Commit} addresses this by generating positive, negative, and uncertain clinical commitments before the report within the same autoregressive trajectory, thereby structuring the clinical evidence that conditions the subsequent report. Overall, \modelname{} progressively coordinates available evidence under varying input configurations. We validate the framework through extensive experiments on MIMIC-RG4 and MIMIC-CXR.

Our main contributions are as follows:
\begin{itemize}
    \item To the best of our knowledge, we are the first to explicitly formulate structured incomplete-context RRG as an availability-conditioned evidence-routing problem, emphasizing how observed evidence should be progressively utilized rather than merely accommodated as source availability changes.
    
    \item We develop \modelname{} to implement multi-interface evidence routing through three coordinated mechanisms: endpoint-preserving patchwise depth routing, availability-conditioned low-rank prefix calibration, and polarity-structured clinical commitment.
    
    \item Experiments on MIMIC-RG4 and MIMIC-CXR show that \modelname{} achieves state-of-the-art CE F1 across all four MIMIC-RG4 settings while maintaining competitive language-generation performance, together with strong conventional-RRG results. Component and mechanism ablations further validate the proposed routing design.
\end{itemize}

\section{Related Work}
\label{sec:related}

We review three lines of work most relevant to our setting: flexible and context-enriched RRG, incomplete multimodal learning, and evidence-grounded structured generation.

\paragraph{Flexible and context-enriched RRG.}
Multi-view and longitudinal RRG exploit additional projections and historical information to enrich spatial and temporal evidence \citep{liu2025hcllm,liu2025mlrg,liu2026priorrg}, while recent methods further model disease evolution through temporal decoupling and progression-aware prompting \citep{dong2026tim,liu2026biotprompt}. LLM-RG4 establishes a flexible four-context setting with adaptive token fusion and token-level weighting for variable inputs \citep{wang2025llmrg4}. These methods improve contextual flexibility and evidence integration under variable input conditions. \modelname{} complements this line by focusing on effective evidence utilization rather than input accommodation alone as source availability changes.

\paragraph{Incomplete multimodal learning.}
Existing methods address missing inputs through reconstruction, representation decoupling, and adaptive fusion \citep{liu2023m3ae,wang2024gmd,yao2024drfuse}, or through dynamic weighting, retrieval, and prompting \citep{li2025simmlm,lang2025ragpt,pipoli2025missrag}. In RRG, DiA-gnostic VLVAE handles missing clinical context through MoE-based shared-latent inference, allowing the shared posterior to rely on observed modalities when context is unavailable \citep{shaik2026diagnostic}. In contrast, \modelname{} focuses on availability-conditioned adaptation of the shared generator, calibrating its language-model prefix according to both observed evidence and source availability.

\paragraph{Evidence grounding and structured generation.}
RRG methods improve visual grounding through hierarchical representations, anatomical regions, and auxiliary alignment \citep{huang2023kiut,tanida2023region,gao2026s2dalign}. Structured generation has explored description planning, diagnosis-derived prompting, observation planning, topic organization, and explicit clinical reasoning \citep{nishino2022coplan,jin2024promptmrg,hou2023organ,cheng2026llavata,zhang2026chexone}. Recent work further explores self-critique and confidence-guided rewriting \citep{yan2026radscr,yu2026anchordiff}. Compared with prior work, \modelname{} couples endpoint-preserving visual refinement with polarity-structured clinical commitments to guide subsequent report generation.

\section{Method}
\label{sec:method}

\subsection{Problem Formulation and Framework Overview}
\label{sec:problem-formulation}

We study radiology report generation under structured incomplete clinical context, where incompleteness is defined at the source level over three predefined sources: a frontal radiograph $\mathbf I_f$, a lateral radiograph $\mathbf I_l$, and a previous report $\mathbf T_p$. The frontal radiograph is always observed, whereas the lateral radiograph and previous report may be unavailable. Indication/history, when available, is included as prompt context and is not treated as an availability axis.

We represent auxiliary-source availability by $\mathbf a=(a_l,a_p)\in\{0,1\}^2$, where $a_l$ and $a_p$ indicate the presence of the lateral view and previous report. SN, SW, MN, and MW correspond to $(0,0)$, $(0,1)$, $(1,0)$, and $(1,1)$, respectively. Here S/M denote frontal-only/frontal--lateral image inputs, and N/W denote the absence/presence of a previous report. Let $\mathbf Y=(y_1,\ldots,y_T)$ be the scenario-specific target report. Following LLM-RG4 \citep{wang2025llmrg4}, unavailable sources occupy zero-valued feature slots in a fixed source layout.

\begin{figure*}[t]
\centering
\includegraphics[width=1\textwidth]{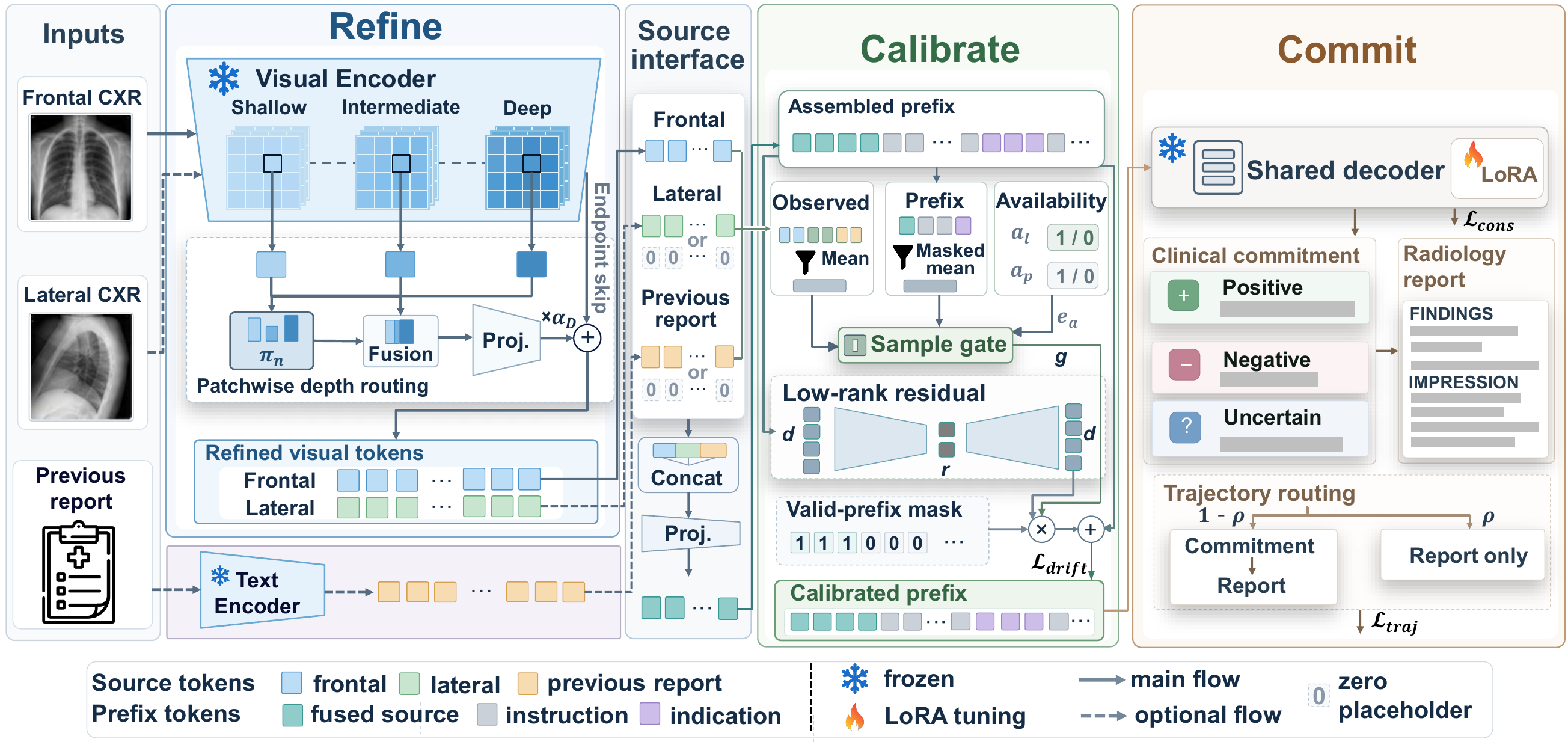}
\caption{Overview of \modelname{} under structured incomplete clinical context. Refine enhances observed visual evidence through endpoint-preserving patchwise depth routing, Calibrate applies availability-conditioned prefix calibration, and Commit generates polarity-structured clinical commitments before the report in the same autoregressive trajectory.}
\label{fig:pacer_overview}
\end{figure*}

\modelname{} progressively routes observed evidence through the \textbf{Refine--Calibrate--Commit} pipeline (Figure~\ref{fig:pacer_overview}). \textbf{Refine} processes the observed radiographs, while a separate text encoder handles the previous report; their representations are assembled into the language-model prefix. \textbf{Calibrate} adapts this prefix to the observed evidence and availability state, and \textbf{Commit} generates polarity-structured commitments to guide report content.

\subsection{Refine: Endpoint-Preserving Patchwise Depth Routing}

Representations from different Transformer depths encode complementary visual information \citep{ranftl2021dpt,lin2025multilayer,huang2023kiut}. Because clinically relevant radiographic evidence can be spatially localized \citep{tanida2023region}, we allow each patch to select its own depth distribution rather than using a global mixture shared across the image.

For each observed radiograph $\mathbf I_s$, $s\in\{f,l\}$, let
$\mathbf V_s^{(\ell)}=[\mathbf v_{s,1}^{(\ell)},\ldots,\mathbf v_{s,N_v}^{(\ell)}]^\top\in\mathbb R^{N_v\times d_v}$
denote the non-CLS patch features extracted by frozen RAD-DINO \citep{perezgarcia2025raddino} at encoder depth $\ell\in\mathcal K$. Here, $N_v$ is the number of non-CLS patch tokens, $d_v$ is their feature dimension, and $\mathcal K$ is the set of selected encoder depths. The same encoder and depth router are applied independently to the frontal and lateral sources. For readability, we omit the source index $s$ below and write $\mathbf V^{(\ell)}$ and $\mathbf v_n^{(\ell)}$. To align depth-dependent feature statistics, we map each patch into a common routing space:
\begin{equation}
\mathbf u_n^{(\ell)}
=
\mathbf P_{\ell}
\opLN_{\ell}\!\left(\mathbf v_n^{(\ell)}\right)
+\mathbf b_{\ell},
\label{eq:depth-alignment}
\end{equation}
where $\mathbf P_{\ell}\in\mathbb R^{d_r\times d_v}$ is a learnable projection, $\opLN_{\ell}$ is a layer-specific normalization, $\mathbf b_{\ell}\in\mathbb R^{d_r}$ is the corresponding bias, and $d_r$ is the routing dimension.

A shared scorer produces a patch-specific distribution over the selected depths:
\begin{equation}
\pi_{n,\ell}
=
\frac{
\exp\!\left(
\mathbf w_{\pi}^{\top}\mathbf u_n^{(\ell)}+\beta_{\ell}
\right)}
{
\sum_{j\in\mathcal K}
\exp\!\left(
\mathbf w_{\pi}^{\top}\mathbf u_n^{(j)}+\beta_j
\right)},
\qquad
\sum_{\ell\in\mathcal K}\pi_{n,\ell}=1,
\label{eq:depth-router}
\end{equation}
where $\mathbf w_{\pi}\in\mathbb R^{d_r}$ is the shared routing vector, $\beta_{\ell}\in\mathbb R$ is a learnable depth-specific bias, and $j$ indexes candidate depths in $\mathcal K$. Thus, $\boldsymbol\pi_n=(\pi_{n,\ell})_{\ell\in\mathcal K}$ varies across patches, allowing spatially localized evidence to draw differently on representations from different encoder depths rather than sharing a single image-level depth mixture.

To incorporate complementary cross-depth cues without replacing the pretrained terminal representation, we use the routed mixture to predict an additive correction. This preserves a direct path from the pretrained endpoint while allowing each patch to incorporate information from earlier encoder depths:
\begin{align}
\boldsymbol\delta_n
&=
\mathbf W_o
\opLN_{\delta}\!\left(
\sum_{\ell\in\mathcal K}
\pi_{n,\ell}\mathbf u_n^{(\ell)}
\right)
+\mathbf b_o,
\\
\widehat{\mathbf v}_n
&=
\mathbf v_n^{\mathrm{end}}
+
\alpha_D\boldsymbol\delta_n,
\qquad
\alpha_D=\bar{\alpha}_D\opsigmoid(\eta_D),
\label{eq:depth-residual}
\end{align}
where $\mathbf v_n^{\mathrm{end}}$ denotes the final normalized RAD-DINO patch representation, $\opLN_{\delta}$ normalizes the routed mixture, $\mathbf W_o\in\mathbb R^{d_v\times d_r}$ and $\mathbf b_o\in\mathbb R^{d_v}$ project it back to the encoder feature space, $\eta_D$ is a learnable scalar, and $\bar{\alpha}_D$ bounds the scalar residual weight $\alpha_D$.

With $\mathbf W_o$ and $\mathbf b_o$ initialized to zero, the module starts exactly from the pretrained terminal representation and learns patch-specific corrections during training. Restoring the source index, Refine outputs $\widehat{\mathbf V}_f$ and, when available, $\widehat{\mathbf V}_l$, which are passed to the source interface used by Calibrate.

\subsection{Calibrate: Availability-Calibrated Prefix Routing}
\label{sec:prefix-routing}

Motivated by context-aware prompting for incomplete multimodal inputs \citep{lang2025ragpt,pipoli2025missrag}, we adapt the pre-Transformer prefix to observed evidence and source availability. Each observed source---the refined image features $\widehat{\mathbf V}_f,\widehat{\mathbf V}_l$ or the text encoding of $\mathbf T_p$---is compressed and projected into pre-fusion tokens $\mathbf Z_s\in\mathbb R^{N_q\times d}$, $s\in\{f,l,p\}$, where $N_q$ is the number of compressed tokens per source and $d$ is the language-model embedding dimension. Unavailable sources use zero-valued slots. The source tokens are fused, then assembled with the prompt context into $\mathbf E_P$.

During teacher forcing, $\mathbf E=[\mathbf E_P;\mathbf E_T]\in\mathbb R^{T'\times d}$ concatenates this prefix with the target-side embeddings $\mathbf E_T$ of the routed sequence in Section~\ref{sec:clinical-commitment}, where $T'$ counts input positions. The mask $\mathbf m\in\{0,1\}^{T'}$ selects valid prefix positions, excluding target-side positions and padding. We define
\begin{equation}
\begin{aligned}
\bar{\mathbf h}_P
&=
\frac{\mathbf E^{\top}\mathbf m}
{\lVert\mathbf m\rVert_1},
\qquad
\bar{\mathbf z}_s
=
\frac{(\mathbf Z_s)^{\top}\mathbf 1_{N_q}}{N_q},
\\
\bar{\mathbf z}_{\mathbf a}
&=
\frac{
\bar{\mathbf z}_f
+a_l\bar{\mathbf z}_l
+a_p\bar{\mathbf z}_p
}{
1+a_l+a_p
},
\end{aligned}
\label{eq:prefix-summaries}
\end{equation}
where $\lVert\mathbf m\rVert_1$ counts valid prefix positions, $\mathbf 1_{N_q}$ is an $N_q$-dimensional all-ones vector, $\bar{\mathbf h}_P\in\mathbb R^d$ is the prefix summary, and $\bar{\mathbf z}_{\mathbf a}\in\mathbb R^d$ averages only observed sources, avoiding dilution by missing-source zero slots.

Each availability state $\mathbf a$ is associated with a learnable embedding $\mathbf e_{\mathbf a}\in\mathbb R^{d_a}$, where $d_a$ is the availability-embedding dimension. The three inputs provide complementary conditioning cues: $\bar{\mathbf h}_P$ summarizes the assembled prefix after source fusion and prompting, $\bar{\mathbf z}_{\mathbf a}$ explicitly summarizes the observed sources before fusion, and $\mathbf e_{\mathbf a}$ identifies which optional sources are available. Their combination conditions the adaptation strength on both evidence content and source availability:
\begin{equation}
g
=
\opsigmoid\!\left(
\mathbf w_g^{\top}
\opGELU\!\left(
\mathbf W_g
[\bar{\mathbf h}_P;
 \bar{\mathbf z}_{\mathbf a};
 \mathbf e_{\mathbf a}]
+\mathbf b_g
\right)
+b_{\mathrm{out}}
\right),
\label{eq:prefix-gate}
\end{equation}
where $[\cdot;\cdot]$ denotes concatenation, $\mathbf W_g\in\mathbb R^{d_g\times(2d+d_a)}$ and $\mathbf b_g\in\mathbb R^{d_g}$ parameterize the hidden layer, $\mathbf w_g\in\mathbb R^{d_g}$ and $b_{\mathrm{out}}\in\mathbb R$ are the output projection and scalar bias, respectively, and $d_g$ is the gate hidden dimension. The resulting scalar $g\in(0,1)$ controls the adaptation strength for each sample.

To provide token-specific corrections with a compact residual parameterization, we use a position-wise bottleneck mapping $\mathcal B:\mathbb R^{T'\times d}\rightarrow\mathbb R^{T'\times d}$, scaled by the sample-wise gate $g$:
\begin{align}
\mathcal B(\mathbf E)
&=
\operatorname{Dropout}\!\left[
\opGELU\!\left(
\opLN(\mathbf E)\mathbf W_{\downarrow}^{\top}
\right)
\right]
\mathbf W_{\uparrow}^{\top},
\notag\\
\widetilde{\mathbf E}
&=
\mathbf E
+
g\,\operatorname{Diag}(\mathbf m)\mathcal B(\mathbf E),
\label{eq:prefix-route}
\end{align}
where $\opLN$ denotes layer normalization, $\mathbf W_{\downarrow}\in\mathbb R^{r\times d}$ and $\mathbf W_{\uparrow}\in\mathbb R^{d\times r}$ are the learnable down- and up-projection matrices, $r\ll d$ is the bottleneck dimension, and $\operatorname{Diag}(\mathbf m)\in\mathbb R^{T'\times T'}$ is the diagonal mask induced by $\mathbf m$. Thus, $\mathcal B(\mathbf E)$ is a token-wise correction with the same shape as $\mathbf E$, while only valid prefix positions receive this correction. Zero-initializing $\mathbf W_{\uparrow}$ gives $\widetilde{\mathbf E}=\mathbf E$ at initialization.

The gate is target-blind because it uses only valid-prefix information, observed sources, and availability, not $\mathbf E_T$. Together with the position-wise $\mathcal B$, this gives $\partial\widetilde{\mathbf E}_P/\partial\mathbf E_T=\mathbf 0$. The mask ensures $[\widetilde{\mathbf E}-\mathbf E]_{t,:}=\mathbf 0$ whenever $m_t=0$. The same prefix-only correction is applied before the Transformer during teacher forcing and autoregressive generation.

\subsection{Commit: Polarity-Structured Clinical Commitment Routing}
\label{sec:clinical-commitment}

Building on structured RRG generation \citep{nishino2022coplan,jin2024promptmrg,hou2023organ}, the shared decoder uses the calibrated prefix $\widetilde{\mathbf E}_P$ to generate a polarity-structured clinical commitment before the report, providing structured clinical context for subsequent generation. Supervision is provided by $\mathbf q^\star=(\mathbf q^{\mathrm{pos}},\mathbf q^{\mathrm{neg}},\mathbf q^{\mathrm{unc}})$, which organizes positive, negative, and uncertain findings, with its construction detailed in Appendix~\ref{app:implementation}.

To combine explicit clinical commitment supervision with direct-report supervision, we stochastically route complete supervised trajectories through the same decoder. Let $\xi\sim\operatorname{Bernoulli}(1-\rho)$, where $\rho$ is the probability of selecting the direct-report route, and let $\boldsymbol\iota_{\mathrm C}$ and $\boldsymbol\iota_{\mathrm D}$ denote the commitment-first and direct-report instructions, respectively. The corresponding route-specific target sequence $\boldsymbol\tau_{\xi}^{\star}$ is
\begin{equation}
(\boldsymbol\iota_{\xi},\boldsymbol\tau_{\xi}^{\star})
=
\begin{cases}
\left(
\boldsymbol\iota_{\mathrm C},
[
\langle\textsc{Anchor}\rangle,
\mathbf q^{\star},
\langle/\textsc{Anchor}\rangle,
\langle\textsc{Report}\rangle,
\mathbf Y,
\langle/\textsc{Report}\rangle
]
\right), & \xi=1,
\\[1mm]
(\boldsymbol\iota_{\mathrm D},\mathbf Y), & \xi=0.
\end{cases}
\label{eq:commitment-route}
\end{equation}
The route is sampled before instruction and target construction, so routing changes the complete supervised sequence rather than isolated tokens. Inference always uses the commitment-first route, with one decoder generating the commitment followed by the report in a single left-to-right pass.

\subsection{Overall Learning Objective}
\label{sec:objective}

Training follows the Refine--Calibrate--Commit design in two optimization stages (phase details in Appendix~\ref{app:implementation}). Refine and Commit are first optimized with autoregressive trajectory supervision; the resulting parent is then frozen and only Calibrate is trained with output- and representation-level regularization.

For training example $b$, let $\xi_b$ denote the route sampled as in Section~\ref{sec:clinical-commitment}, and let $\boldsymbol\tau_b^\star=\boldsymbol\tau_{\xi_b}^\star$ denote the corresponding routed target sequence. We then let $\Omega_{\mathrm Q}^{(b)}$ and $\Omega_{\mathrm Y}^{(b)}$ denote its supervised commitment- and report-side positions, respectively, with $\Omega_{\mathrm Q}^{(b)}=\varnothing$ for direct-report samples, and define $\mathcal S_b=\Omega_{\mathrm Q}^{(b)}\cup\Omega_{\mathrm Y}^{(b)}$. Given
\[
\ell_{bt}
=
-\log p_\theta
\left(
\tau_{bt}^{\star}
\mid
\boldsymbol\tau_{b,<t}^{\star},
\mathbf E_{P,b}
\right),
\]
where $p_\theta$ denotes the complete model; observed-source and availability conditioning is omitted for brevity. To balance commitment and report supervision, we set $\omega_{bt}=\lambda_{\mathrm Q}$ on $\Omega_{\mathrm Q}^{(b)}$ and $\omega_{bt}=1$ on $\Omega_{\mathrm Y}^{(b)}$. Let $\kappa_b=\lambda_{\mathrm Q}$ for commitment-first samples and $\kappa_b=1$ for direct-report samples denote the ignored-BOS weight retained by the implementation normalizer. The trajectory objective is
\begin{equation}
\mathcal L_{\mathrm{traj}}
=
\frac{
\sum_b\sum_{t\in\mathcal S_b}\omega_{bt}\ell_{bt}
}{
\sum_b\sum_{t\in\mathcal S_b}\omega_{bt}
+\sum_b\kappa_b+\varepsilon
}
+
\lambda_{\mathrm{TLW}}\mathcal L_{\mathrm{TLW}},
\label{eq:trajectory-objective}
\end{equation}
where $\varepsilon$ is a numerical stabilizer and $\mathcal{L}_{\mathrm{TLW}}$ is an independently normalized report-side auxiliary loss using the emphasized-token annotations from LLM-RG4 \citep{wang2025llmrg4}. Exact span boundaries, masking, and token weighting are detailed in Appendix~\ref{app:implementation}.

During Calibrate training, the frozen parent provides a reference for limiting output deviation and prefix drift. Let $p_0$ denote the next-token distribution of the frozen parent model with prefix routing disabled, and $p_\theta$ that of the adapted model, both conditioned on the same observed case. Let $\Omega_{\mathrm{cons}}^{(b)}$ denote the valid prediction positions used for output consistency. We regularize output deviation and valid-prefix representation drift by
\begin{align}
\mathcal L_{\mathrm{cons}}
&=
\mathbb E_{b,t\in\Omega_{\mathrm{cons}}^{(b)}}
\opKL\!\left(
p_0(\cdot\mid\boldsymbol\tau_{b,<t}^{\star},\mathbf E_{P,b})
\Vert
p_\theta(\cdot\mid\boldsymbol\tau_{b,<t}^{\star},\mathbf E_{P,b})
\right),
\label{eq:parent-consistency}
\\
\mathcal L_{\mathrm{drift}}
&=
\mathbb E_{b,t:m_{bt}=1}
\frac{
\left\lVert
\widetilde{\mathbf E}_{bt,:}-\mathbf E_{bt,:}
\right\rVert_2^2
}{d},
\label{eq:prefix-drift}
\end{align}
where $\mathbf m_b$ is the prefix mask defined in Section~\ref{sec:prefix-routing}. The Calibrate stage minimizes
\begin{equation}
\mathcal L_{\mathrm{PACER}}
=
\mathcal L_{\mathrm{traj}}
+
\lambda_{\mathrm{cons}}\mathcal L_{\mathrm{cons}}
+
\lambda_{\mathrm{drift}}\mathcal L_{\mathrm{drift}},
\label{eq:complete-objective}
\end{equation}
where $\lambda_{\mathrm{cons}}$ and $\lambda_{\mathrm{drift}}$ weight the output-consistency and prefix-drift regularizers, respectively.

\begin{table*}[t]
\centering
\caption{Four-context results on MIMIC-RG4. Published baselines follow \citet{wang2025llmrg4}; $\ast$ denotes CXRMate retrained therein and $\dagger$ our SimMLM/RAGPT adaptations. Bold/underline indicate the best/second-best non-tied values per context; ties for best are bold.}
\label{tab:main_rg4}
\begingroup
\setlength{\tabcolsep}{2.8pt}
\begin{tabular}{@{}llccccccccc@{}}
\toprule
\multirow{2}{*}{Setting}
& \multirow{2}{*}{Model}
& \multicolumn{3}{c}{CE Metrics}
& \multicolumn{6}{c}{NLG Metrics} \\
\cmidrule(lr){3-5}
\cmidrule(lr){6-11}
& & P & R & F1
& B@1 & B@2 & B@3 & B@4 & R-L & MTR \\
\midrule

\multirow{6}{*}{SN}
& CXRMate$^{\ast}$
& 0.572 & 0.560 & 0.566
& 0.421 & 0.271 & 0.179 & 0.122 & 0.311 & 0.174 \\

& RadFM
& 0.413 & 0.303 & 0.350
& 0.188 & 0.090 & 0.048 & 0.028 & 0.190 & 0.094 \\

& LLM-RG4
& \underline{0.588} & \underline{0.632} & \underline{0.609}
& \textbf{0.479} & \textbf{0.343} & \textbf{0.255}
& \textbf{0.196} & \underline{0.384} & \textbf{0.209} \\

& SimMLM$^{\dagger}$
& 0.568 & 0.591 & 0.579
& 0.425 & 0.277 & 0.186 & 0.128 & 0.313 & 0.168 \\

& RAGPT$^{\dagger}$
& 0.565 & 0.564 & 0.564
& 0.423 & 0.278 & 0.188 & 0.129 & 0.314 & 0.168 \\

& \textbf{PACER (Ours)}
& \textbf{0.620} & \textbf{0.645} & \textbf{0.632}
& \underline{0.468} & \underline{0.337} & \underline{0.251}
& \underline{0.192} & \textbf{0.390} & \underline{0.208} \\

\midrule

\multirow{6}{*}{SW}
& CXRMate$^{\ast}$
& 0.573 & 0.549 & 0.561
& 0.361 & 0.220 & 0.139 & 0.093 & 0.284 & 0.153 \\

& RadFM
& 0.508 & 0.365 & 0.425
& 0.211 & 0.103 & 0.056 & 0.033 & 0.183 & 0.105 \\

& LLM-RG4
& \underline{0.599} & \underline{0.622} & \underline{0.610}
& \underline{0.455} & \underline{0.321} & \underline{0.239}
& \textbf{0.186} & \underline{0.382} & \underline{0.199} \\

& SimMLM$^{\dagger}$
& 0.583 & 0.556 & 0.569
& 0.379 & 0.236 & 0.152 & 0.105 & 0.306 & 0.152 \\

& RAGPT$^{\dagger}$
& 0.568 & 0.556 & 0.562
& 0.387 & 0.244 & 0.158
& \underline{0.106} & 0.307 & 0.156 \\

& \textbf{PACER (Ours)}
& \textbf{0.623} & \textbf{0.654} & \textbf{0.638}
& \textbf{0.456} & \textbf{0.323} & \textbf{0.240}
& \textbf{0.186} & \textbf{0.386} & \textbf{0.202} \\

\midrule

\multirow{6}{*}{MN}
& CXRMate$^{\ast}$
& \underline{0.544} & 0.522 & 0.533
& 0.437 & 0.289 & 0.199 & 0.141 & 0.332 & 0.179 \\

& RadFM
& 0.323 & 0.187 & 0.237
& 0.246 & 0.113 & 0.060 & 0.034 & 0.194 & 0.104 \\

& LLM-RG4
& 0.541 & \underline{0.578} & \underline{0.559}
& \textbf{0.491} & \textbf{0.359} & \textbf{0.274}
& \textbf{0.216} & \underline{0.405} & \underline{0.214} \\

& SimMLM$^{\dagger}$
& 0.512 & 0.558 & 0.534
& 0.429 & 0.283 & 0.195 & 0.138 & 0.325 & 0.172 \\

& RAGPT$^{\dagger}$
& \underline{0.544} & 0.531 & 0.537
& 0.431 & 0.288 & 0.200 & \underline{0.142} & 0.327 & 0.174 \\

& \textbf{PACER (Ours)}
& \textbf{0.586} & \textbf{0.580} & \textbf{0.583}
& \underline{0.483} & \underline{0.357} & \underline{0.273}
& \textbf{0.216} & \textbf{0.413} & \textbf{0.215} \\

\midrule

\multirow{6}{*}{MW}
& CXRMate$^{\ast}$
& 0.548 & 0.499 & 0.523
& 0.379 & 0.241 & 0.158 & 0.110 & 0.305 & 0.159 \\

& RadFM
& 0.456 & 0.297 & 0.360
& 0.191 & 0.095 & 0.054 & 0.034 & 0.178 & 0.095 \\

& LLM-RG4
& \underline{0.560} & \underline{0.565} & \underline{0.563}
& \textbf{0.461} & \textbf{0.331} & \textbf{0.250}
& \underline{0.197} & \underline{0.401} & \underline{0.204} \\

& SimMLM$^{\dagger}$
& 0.529 & 0.538 & 0.533
& 0.387 & 0.245 & 0.161 & 0.113 & 0.317 & 0.157 \\

& RAGPT$^{\dagger}$
& 0.529 & 0.523 & 0.526
& 0.397 & 0.255 & 0.169 & 0.118 & 0.324 & 0.163 \\

& \textbf{PACER (Ours)}
& \textbf{0.582} & \textbf{0.591} & \textbf{0.586}
& \underline{0.458} & \underline{0.330} & \underline{0.249}
& \textbf{0.198} & \textbf{0.406} & \textbf{0.205} \\

\bottomrule
\end{tabular}
\endgroup
\end{table*}

\section{Experiments}
\label{sec:experiments}

\subsection{Experimental Setup}
\label{sec:setup}

\paragraph{Task and data.}
We evaluate \modelname{} under both structured incomplete-context and conventional RRG settings. The primary evaluation uses MIMIC-RG4 \citep{wang2025llmrg4}, constructed from MIMIC-CXR \citep{johnson2019mimiccxr}, under the four availability states defined in Section~\ref{sec:problem-formulation}. Following the benchmark protocol, the four-context evaluation uses findings-and-impression reports, whereas conventional SN uses findings only. For this fixed-availability setting, we evaluate a dedicated Refine+Commit model, while Calibrate is evaluated in the multi-context setting where structured source availability varies.

\paragraph{Baselines and metrics.}
For the four-context comparison, we use the published LLM-RG4, CXRMate, and RadFM results from \citet{wang2025llmrg4}, and include local adaptations of SimMLM \citep{li2025simmlm} and RAGPT \citep{lang2025ragpt} under the same four-context MIMIC-RG4 evaluation protocol; provenance and adaptation details are provided in Appendix~\ref{app:baseline-adaptation}. Clinical efficacy (CE) is measured by micro-averaged precision, recall, and F1 from CheXbert-extracted labels \citep{smit2020chexbert}, and language quality by BLEU-1--4 (B@1--4), ROUGE-L (R-L), and METEOR (MTR) \citep{papineni2002bleu,lin2004rouge,banerjee2005meteor}. Ablations are summarized by the arithmetic mean over the four MIMIC-RG4 contexts, computed before rounding.

\paragraph{Implementation.}
Following LLM-RG4, we use Vicuna-7B v1.5 with LoRA \citep{hu2022lora} as the language decoder and frozen RAD-DINO as the visual backbone. The depth router uses encoder layers $\{4,8,12\}$, and the prefix residual branch uses rank $r=16$. Clinical commitment targets are constructed offline from reference reports using a hybrid LLM-assisted and rule-based extraction pipeline. Training follows the staged optimization in Section~\ref{sec:objective}; stage-specific trajectory sampling, loss weights, and other implementation details are provided in Appendix~\ref{app:implementation}.

\begin{table*}[t]
\centering
\caption{Conventional findings-only SN results. The upper block provides literature context under original protocols, while the lower block evaluates methods on a common findings-only SN test set, with clean and original references reported separately. $\dagger$ denotes local downstream training initialized from released pretrained weights, and $\ddagger$ denotes a released checkpoint evaluated by us. Bold/underline indicate best/second-best values within the common-test block.}
\label{tab:sn_findings}
\begingroup
\small
\setlength{\tabcolsep}{1.8pt}
\begin{tabular}{@{}lccccccccc@{}}
\toprule
\multirow{2}{*}{Model}
& \multicolumn{3}{c}{CE Metrics}
& \multicolumn{3}{c}{Clean NLG}
& \multicolumn{3}{c}{Original NLG} \\
\cmidrule(lr){2-4}
\cmidrule(lr){5-7}
\cmidrule(lr){8-10}
& P & R & F1
& B@1 & B@4 & R-L
& B@1 & B@4 & R-L \\
\midrule

\multicolumn{10}{@{}l}{\emph{Literature-reported results under original protocols}} \\

KiUT~\citep{huang2023kiut}
& 0.371 & 0.318 & 0.321
& -- & -- & --
& 0.393 & 0.113 & 0.285 \\

RGRG~\citep{tanida2023region}
& 0.461 & 0.475 & 0.447
& -- & -- & --
& 0.373 & 0.126 & 0.264 \\

EKAGen~\citep{bu2024ekagen}
& 0.517 & 0.483 & 0.499
& -- & -- & --
& 0.419 & 0.119 & 0.287 \\

MAIRA-1 (7B)~\citep{hyland2023maira1}
& -- & -- & 0.553
& -- & -- & --
& 0.392 & 0.142 & 0.289 \\

Med-PaLM M (562B)~\citep{tu2024medpalm}
& -- & -- & 0.516
& -- & -- & --
& 0.317 & 0.115 & 0.275 \\

R2-LLM (14.2B)~\citep{liu2024bootstrap}
& 0.465 & 0.482 & 0.473
& -- & -- & --
& 0.402 & 0.128 & 0.291 \\

InVERGe (7B)~\citep{deria2024inverge}
& -- & -- & --
& -- & -- & --
& 0.425 & 0.100 & 0.309 \\

REVTAF~\citep{zhou2025revtaf}
& 0.628 & 0.613 & 0.592
& -- & -- & --
& 0.465 & 0.182 & 0.336 \\

ESC-RL~\citep{zhou2026escrl}
& 0.632 & 0.625 & 0.608
& -- & -- & --
& 0.487 & 0.199 & 0.352 \\

S2D-Align~\citep{gao2026s2dalign}
& 0.613 & 0.606 & 0.608
& -- & -- & --
& 0.422 & 0.149 & 0.332 \\

\midrule

\multicolumn{10}{@{}l}{\emph{Evaluation on the common MIMIC-RG4 test set}} \\

R2Gen~\citep{chen2020r2gen}
& 0.456 & 0.306 & 0.366
& 0.363 & 0.090 & 0.269
& 0.356 & 0.097 & 0.267 \\

R2GenCMN~\citep{chen2021cmn}
& 0.486 & 0.400 & 0.439
& 0.385 & 0.102 & 0.278
& 0.349 & 0.094 & 0.270 \\

CvT2DistilGPT2~\citep{nicolson2023cvt2distilgpt2}
& 0.498 & 0.414 & 0.452
& 0.374 & 0.103 & 0.272
& \underline{0.390} & 0.123 & 0.282 \\

PromptMRG~\citep{jin2024promptmrg}
& \textbf{0.618} & 0.491 & 0.548
& 0.326 & 0.080 & 0.261
& 0.381 & 0.096 & 0.258 \\

R2GenGPT (7B)~\citep{wang2023r2gengpt}
& 0.506 & 0.414 & 0.456
& 0.401 & 0.118 & 0.277
& \textbf{0.396} & 0.113 & 0.273 \\

CheXagent (7B)~\citep{chen2024chexagent}
& 0.506 & 0.306 & 0.381
& 0.265 & 0.058 & 0.239
& 0.189 & 0.040 & 0.208 \\

LLM-RG4~\citep{wang2025llmrg4}
& 0.583 & \underline{0.593} & 0.588
& \underline{0.498} & \underline{0.203} & \textbf{0.387}
& 0.377 & \underline{0.144} & \textbf{0.318} \\

MambaXray-VL-Large$^{\dagger}$~\citep{wang2025cxpmrg}
& 0.555 & 0.526 & 0.540
& 0.449 & 0.137 & \underline{0.328}
& 0.376 & 0.103 & 0.264 \\

CheXOne$^{\ddagger}$~\citep{zhang2026chexone}
& \underline{0.613} & 0.574 & \underline{0.593}
& 0.428 & 0.124 & 0.311
& 0.199 & 0.037 & 0.177 \\

\textbf{PACER (Ours)}
& 0.590
& \textbf{0.654}
& \textbf{0.620}
& \textbf{0.511}
& \textbf{0.208}
& \textbf{0.387}
& 0.389
& \textbf{0.146}
& \underline{0.315} \\

\bottomrule
\end{tabular}
\endgroup
\end{table*}

\subsection{Main Results under Structured Incomplete Contexts}
\label{sec:main-results}

Table~\ref{tab:main_rg4} reports the four-context comparison. \modelname{} achieves the highest CE F1 in all four settings, improving over LLM-RG4 by 0.023--0.028, with gains in both precision and recall. These clinical gains largely preserve language-generation quality: BLEU changes marginally, while ROUGE-L improves in all four settings and METEOR in most. Overall, \modelname{} strengthens clinical-label agreement without materially degrading reference-based language quality.

\begin{table}[t]
\centering
\caption{Component ablations averaged over the four MIMIC-RG4 contexts. Base denotes the local ablation baseline. Bold/underline indicate the best/second-best values.}
\label{tab:core_ablation}
\begin{tabular}{@{}lrrrrrr@{}}
\toprule
Model & P & R & F1 & B@1 & B@4 & R-L \\
\midrule
Base & 0.576 & 0.587 & 0.581 & 0.461 & \underline{0.198} & 0.395 \\
Refine only & 0.575 & 0.598 & 0.586 & \textbf{0.470} & \textbf{0.200} & \underline{0.396} \\
Commit only & \underline{0.599} & 0.587 & 0.593 & 0.455 & 0.194 & \textbf{0.399} \\
Refine + Commit & 0.598 & \underline{0.603} & \underline{0.600} & 0.452 & 0.192 & \underline{0.396} \\
\textbf{PACER (+ Calibrate)} & \textbf{0.603} & \textbf{0.617} & \textbf{0.610} & \underline{0.466} & \underline{0.198} & \textbf{0.399} \\
\bottomrule
\end{tabular}
\end{table}

\begin{table}[t]
\centering
\caption{Mechanism ablations averaged over the four MIMIC-RG4 contexts. For prefix routing, $h$, $z$, and $e$ denote the prefix summary, observed-evidence summary, and availability embedding. Bold/underline indicate best/second-best values within each group.}
\label{tab:mechanism_ablation}
\begingroup
\small
\setlength{\tabcolsep}{3pt}
\renewcommand{\arraystretch}{1.08}
\begin{tabular*}{\linewidth}{@{\extracolsep{\fill}}llrrrrrr@{}}
\toprule
Mechanism & Variant & P & R & F1 & B@1 & B@4 & R-L \\
\midrule
\multirow{3}{*}{\textit{Refine}}
& No Refine (endpoint only) & \textbf{0.599} & \underline{0.587} & 0.593 & \textbf{0.455} & \textbf{0.194} & \textbf{0.399} \\
& Global depth weighting & 0.588 & \textbf{0.603} & \underline{0.595} & 0.444 & 0.185 & 0.390 \\
& Patchwise depth routing & \underline{0.598} & \textbf{0.603} & \textbf{0.600} & \underline{0.452} & \underline{0.192} & \underline{0.396} \\
\midrule
\multirow{3}{*}{\textit{Commit}}
& No Commit (direct report) & 0.575 & \underline{0.598} & 0.586 & \textbf{0.470} & \textbf{0.200} & \textbf{0.396} \\
& Always commitment-first & \underline{0.594} & 0.596 & \underline{0.595} & 0.448 & 0.190 & \underline{0.394} \\
& Stochastic trajectory routing & \textbf{0.598} & \textbf{0.603} & \textbf{0.600} & \underline{0.452} & \underline{0.192} & \textbf{0.396} \\
\midrule
\multirow{5}{*}{\textit{Calibrate}}
& No Calibrate (no prefix routing) & 0.598 & 0.603 & 0.600 & 0.452 & 0.192 & 0.396 \\
& $g(h)$ & 0.601 & 0.611 & 0.606 & 0.464 & 0.196 & \underline{0.398} \\
& $g(h,e)$ & \underline{0.602} & 0.612 & \underline{0.607} & \underline{0.465} & \underline{0.197} & \underline{0.398} \\
& $g(h,z)$ & 0.601 & \underline{0.613} & \underline{0.607} & 0.463 & 0.196 & \underline{0.398} \\
& $g(h,z,e)$ (\modelname{}) & \textbf{0.603} & \textbf{0.617} & \textbf{0.610} & \textbf{0.466} & \textbf{0.198} & \textbf{0.399} \\
\bottomrule
\end{tabular*}
\par\smallskip
\begin{minipage}{\linewidth}
\footnotesize
\textit{Controls.} Refine variants share the Commit training protocol, and Commit variants share the Refine architecture and the checkpoint obtained after the SN Refine warm-up, with Calibrate disabled in both; Calibrate variants share the same frozen Refine+Commit parent.
\end{minipage}
\endgroup
\end{table}

\subsection{Conventional SN Evaluation}
\label{sec:sn}

Table~\ref{tab:sn_findings} further evaluates \modelname{} under the conventional findings-only SN setting. On the common MIMIC-RG4 test set, \modelname{} achieves the highest CE F1 (0.620) and recall (0.654), exceeding LLM-RG4 by 0.032 and 0.061, respectively, while PromptMRG retains the highest precision. For language generation, \modelname{} obtains the best BLEU-1 and BLEU-4 on clean references and ties LLM-RG4 for the best clean-reference ROUGE-L. Against the corresponding original MIMIC-CXR reports, it achieves the best BLEU-4 and the second-best ROUGE-L while remaining competitive on BLEU-1. Literature-reported results under different original protocols are included in the upper block of Table~\ref{tab:sn_findings} for broader context rather than direct ranking.

\subsection{Component and Mechanism Ablations}
\label{sec:ablations}

\paragraph{Component contributions.}
Table~\ref{tab:core_ablation} reveals complementary effects across the three components. Refine primarily improves recall and BLEU overlap, whereas Commit yields a larger gain in precision. Combining them improves both precision and recall over the Base, raising CE F1 from 0.581 to 0.600. Training Calibrate on the frozen Refine+Commit parent further increases F1 to 0.610 while improving the B@1 and B@4 scores of the combined model, supporting complementary roles across the three components.

\paragraph{Mechanism analysis.}
Table~\ref{tab:mechanism_ablation} further clarifies the role of each design. Global depth weighting improves recall but reduces precision and language overlap, whereas patchwise routing preserves the recall gain while recovering precision, yielding the best CE F1 among the depth variants. For commitment routing, always using the commitment-first trajectory improves precision at the cost of BLEU, while stochastic routing further improves F1 and partially restores language overlap. For prefix routing, Calibrate improves over the no-routing control. Under matched gate architectures and parameter counts, explicitly incorporating availability and observed-evidence cues provides additional gains over prefix-only calibration, with their joint use performing best. This supports conditioning adaptation that accounts for both source availability and observed evidence.

\subsection{Qualitative Analysis}
\label{sec:qualitative}

\begin{figure*}[t]
\centering
\noindent

\begin{minipage}[t]{0.17\textwidth}
\vspace{0pt}
\centering

\includegraphics[width=0.88\linewidth]{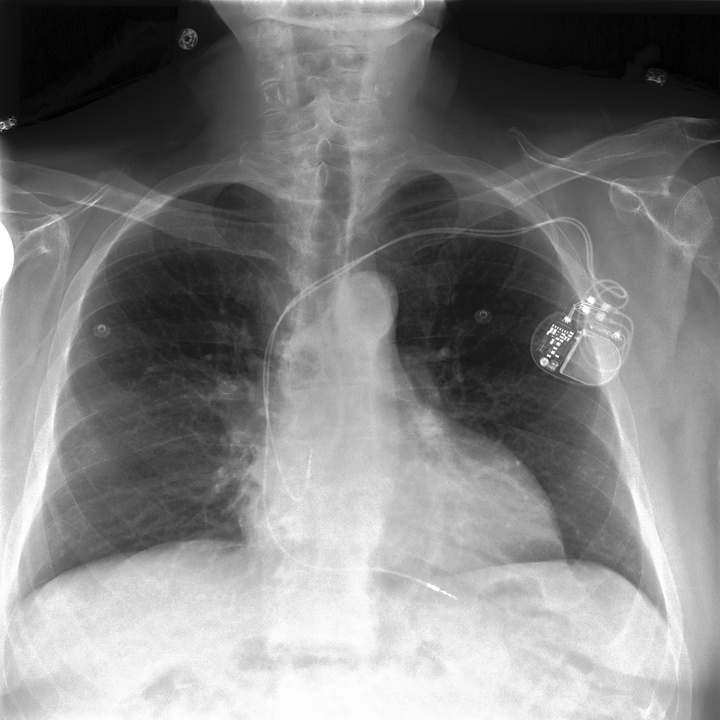}

\vspace{0.12em}

{\footnotesize\textbf{Current frontal CXR}\par}

\vspace{0.32em}

{\footnotesize\textbf{Previous report (SW only)}\par}

\vspace{0.12em}

\raggedright
{\fontsize{8}{6.85}\selectfont
\setlength{\parskip}{0pt}
Dual-chamber pacemaker leads are in appropriate position in the right atrium and right ventricle. No pneumothorax, mediastinal widening, or hemothorax.
\par
}

\end{minipage}
\hfill
\begin{minipage}[t]{0.81\textwidth}
\vspace{0pt}

\setlength{\tabcolsep}{3.8pt}
\renewcommand{\arraystretch}{1.00}

{\fontsize{8}{7.55}\selectfont
\begin{tabularx}{\linewidth}{
    >{\centering\arraybackslash}p{0.095\linewidth}
    >{\RaggedRight\arraybackslash}X
    >{\RaggedRight\arraybackslash}X
    >{\RaggedRight\arraybackslash}X
}
\toprule
\textbf{Context}
&
\textbf{Reference}
&
\textbf{LLM-RG4}
&
\textcolor{CaseBlueText}{\textbf{\modelname{}}}
\\
\midrule

\textbf{SN: no previous report}
&
The right PICC line \matchtxt{projects over the mid SVC}. \matchtxt{The course is unremarkable}. \matchtxt{There is no evidence of complication, notably no pneumothorax}.
&
The PICC tip \matchtxt{projects over the mid-to-low SVC}. \matchtxt{There is no evidence of complications, notably no pneumothorax}. The pacemaker leads are in position.
&
The patient has a right-sided PICC line. \matchtxt{The course is unremarkable and the tip projects over the mid SVC}. \matchtxt{No evidence of complications, notably no pneumothorax}.
\\

\midrule

\textbf{SW: with previous report}
&
As compared with the previous radiograph, \longtxt{a new right PICC line has been inserted}. \matchtxt{The tip projects over the mid SVC}. \matchtxt{The course is unremarkable and there is no complication}.
&
As compared with the previous radiograph, the PICC line has been \unverifiedtxt{pulled back}. \matchtxt{The tip now projects over the mid SVC}. \matchtxt{There is no evidence of complications, notably no pneumothorax}.
&
As compared with the previous radiograph, \pacertxt{the patient has received a right-sided PICC line}. \matchtxt{The course is unremarkable and the tip projects over the mid SVC}. \matchtxt{There is no evidence of complications, notably no pneumothorax}.
\\

\bottomrule
\end{tabularx}
}

\vspace{0.18em}

{\fontsize{8}{7.15}\selectfont
\textbf{Color coding:}
\matchtxt{reference-supported shared content};
\longtxt{reference longitudinal event};
\pacertxt{\modelname{}-captured longitudinal event};
\unverifiedtxt{reference-unverified description}.
\par
\textbf{Longitudinal change:}
the reference indicates a new PICC insertion; \modelname{} captures the same event, whereas LLM-RG4 describes the line as having been pulled back.
}

\end{minipage}

\caption{Qualitative comparison of longitudinal evidence utilization, showing how the generated report changes when the previous report becomes available.}
\label{fig:case_study}
\end{figure*}

Figure~\ref{fig:case_study} illustrates an example of longitudinal evidence utilization. Without the previous report, both \modelname{} and LLM-RG4 largely agree on the current PICC position and the absence of complications. The key difference emerges when the previous report becomes available: \modelname{} introduces the reference-supported new PICC insertion, whereas the fixed LLM-RG4 checkpoint describes the line as having been pulled back. Meanwhile, the current-image findings remain largely consistent across the two contexts. This example suggests that \modelname{} can adapt the longitudinal interpretation of the report to newly available previous-report evidence while preserving the description of current findings. Further discussion of the findings and study limitations is provided in Appendix~\ref{app:discussion}.

\section{Conclusion}
\label{sec:conclusion}

In this work, we presented \modelname{}, a Progressive Availability-Conditioned Evidence Routing framework for radiology report generation under structured incomplete clinical context. Its Refine--Calibrate--Commit pipeline coordinates available evidence across visual representation, language-model conditioning, and clinical report generation within a unified model. \modelname{} achieves state-of-the-art CE F1 across all four MIMIC-RG4 settings while maintaining competitive language-generation performance, with strong results in the conventional MIMIC-CXR setting. These findings highlight effective evidence utilization, beyond accommodating variable inputs alone, as an important design dimension for flexible-context RRG.

\clearpage
\section*{AI Use Statement}
Generative AI tools were used to assist with literature discovery, research ideation and methodological or experimental feedback, manuscript organization, language editing, LaTeX preparation, and consistency checking. All AI-assisted suggestions, reported experimental results, scientific claims, citations, and implementation descriptions were reviewed and verified by the authors. Separately, a hybrid LLM-assisted and rule-based extraction pipeline was used to construct offline clinical commitment supervision, as described in Appendix~A; this external pipeline is not required during inference. The authors take responsibility for the final content of this work.

\section*{Reproducibility Statement}
Section~\ref{sec:method} specifies the proposed routing mechanisms, information-flow constraints, staged optimization, and learning objectives. Appendix~\ref{app:implementation} provides implementation details including module initialization, prefix masking, commitment supervision, trajectory routing, report extraction, and loss configuration. Tables~\ref{tab:main_rg4} and~\ref{tab:sn_findings} report the complete main comparisons, while Appendices~\ref{app:evaluation-notes}
and~\ref{app:baseline-adaptation} document evaluation protocols, result provenance, local baseline adaptations, and ablation settings. Together, these sections specify the methodological and evaluation procedures used for the reported experiments.

\bibliographystyle{iclr2027_conference}
\bibliography{iclr2027_conference}

\clearpage
\raggedbottom
\section*{Author Biographies}

\newcommand{\pacerbio}[3]{%
  \par\noindent
  \begin{minipage}[t]{0.16\textwidth}
    \vspace{0pt}
    \centering
    \includegraphics[width=\linewidth]{#1}
  \end{minipage}
  \hfill
  \begin{minipage}[t]{0.80\textwidth}
    \vspace{0pt}
    \small
    \textbf{#2} #3
  \end{minipage}
  \par\vspace{2pt}
}

\pacerbio{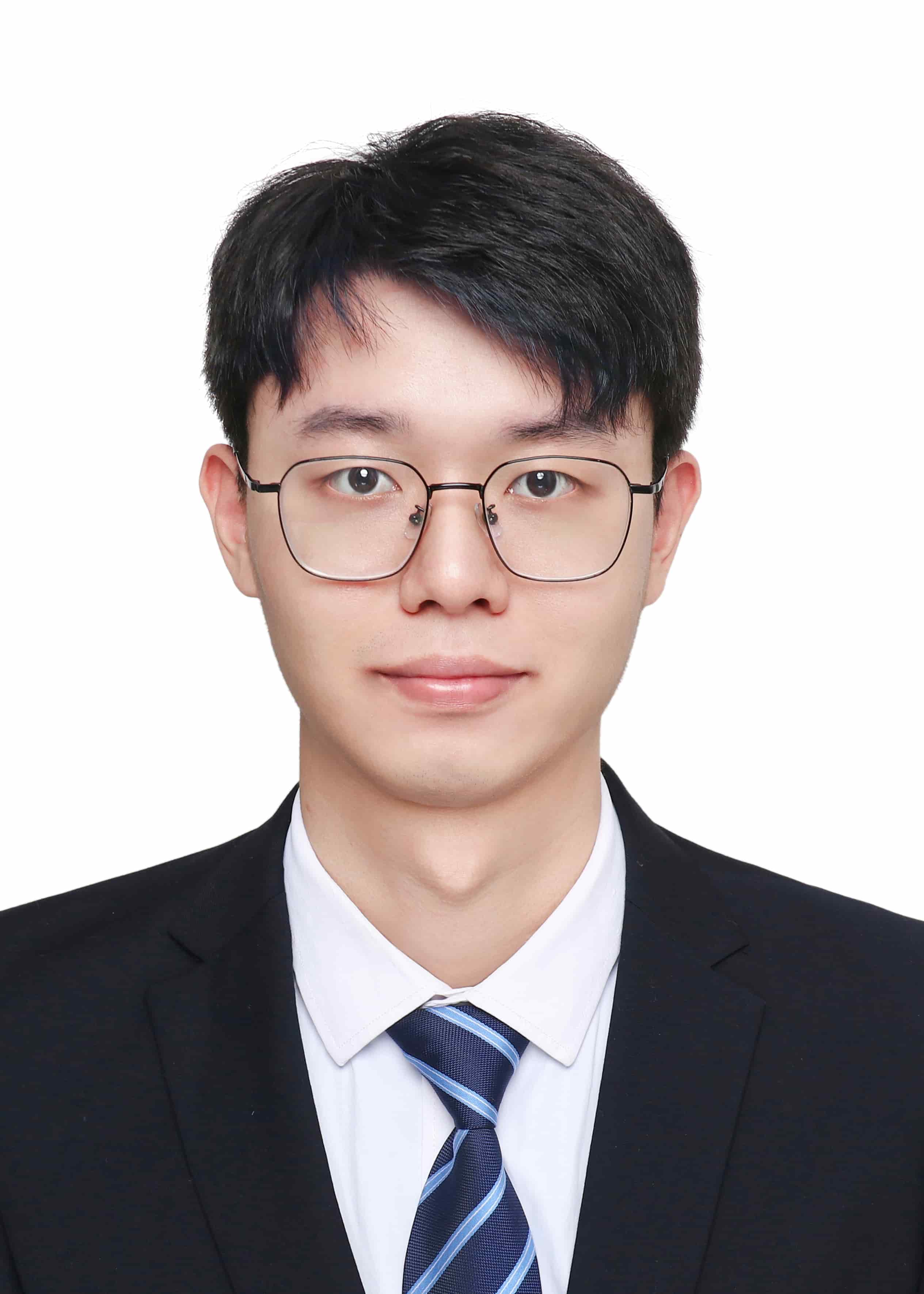}{Yulong Chen}{
received the B.S. degree in Computer Science and Technology from South China University of Technology in 2023. He is currently pursuing the Ph.D. degree with the School of Computer Science and Technology, Harbin Institute of Technology, Shenzhen, China. His research interests include the safety of large language models against jailbreak attacks and multimodal large language models in the medical domain.
}

\pacerbio{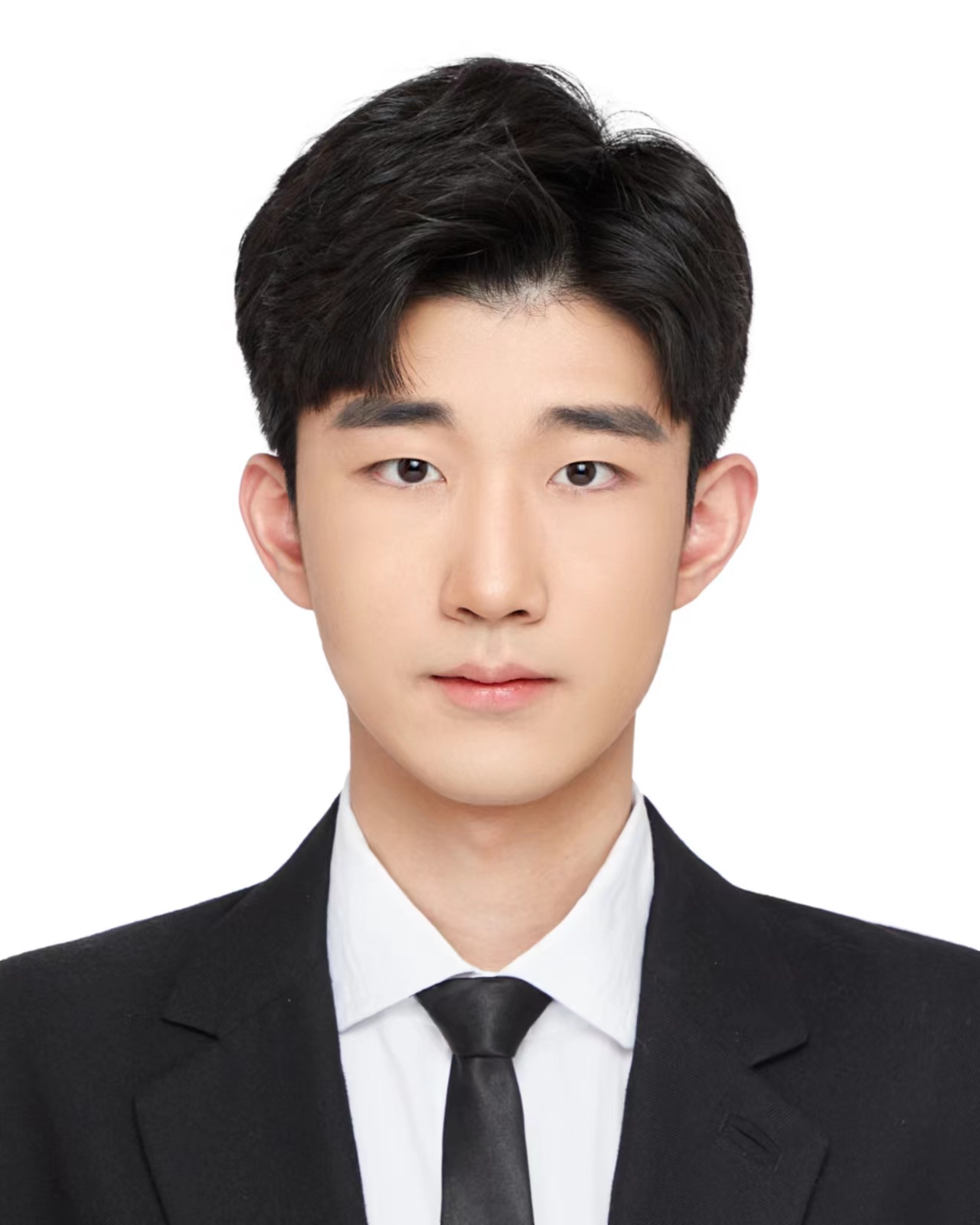}{Yadong Liu}{
(Student Member, IEEE) received the M.S. degree in computer science from The Chinese University of Hong Kong, Hong Kong, China, in 2024, and the B.S. degree in cyberspace security from Harbin Institute of Technology, Weihai, China, in 2023. He is currently a Ph.D. student in the School of Computer Science and Technology of Harbin Institute of Technology, Shenzhen, China. 
}

\pacerbio{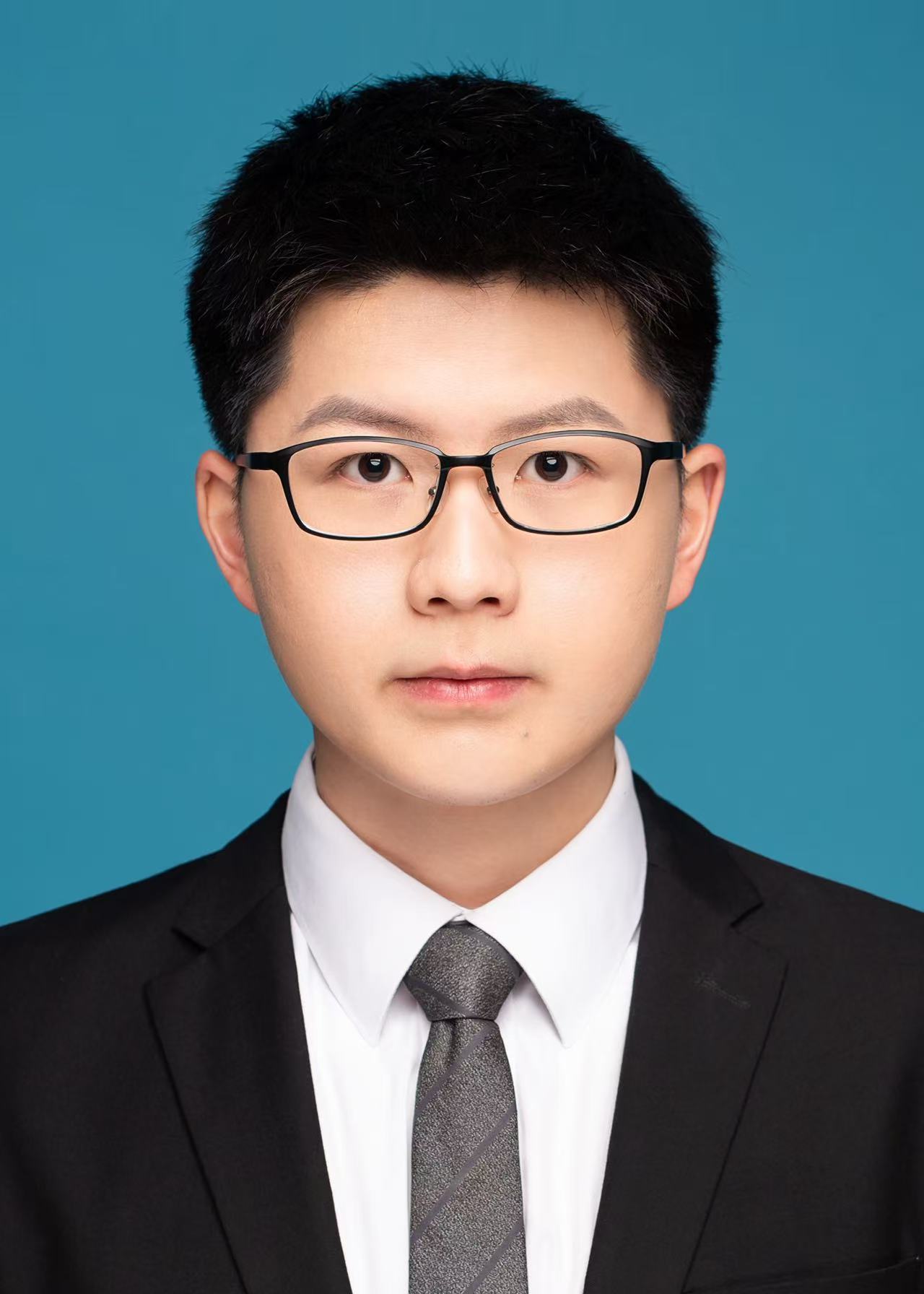}{Haoyu Cao}{
received his B.S. degree from the School of Science, Harbin Institute of Technology, in 2022, and the M.S. degree  from the School of Science, Harbin Institute of Technology, Shenzhen, in 2025. He is currently pursuing his Ph.D. degree in Computer Science and Technology at Harbin Institute of Technology, Shenzhen. His research interests include medical image processing, multimodal diagnosis, and agentic diagnosis.
}

\pacerbio{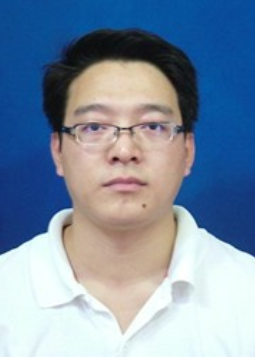}{Sen Xu}{
was born in Yancheng, China, in 1983. He received B.S., M.S., and Ph.D. degrees in computer science from Harbin Engineering University, Harbin, China, in 2004, 2007, and 2010, respectively. He joined Yancheng Institute of Technology, Yancheng, China, in 2010, where he is currently a Professor with the School of Information Engineering. His research interests include pattern recognition, machine learning, and data mining. His current research focuses on cluster ensemble and multiview clustering. Dr. Xu is a reviewer for several high-quality international journals, including Artificial Intelligence Review, Pattern Recognition, and Neurocomputing. He is also a member of China Computer Federation and CAAI.
}

\pacerbio{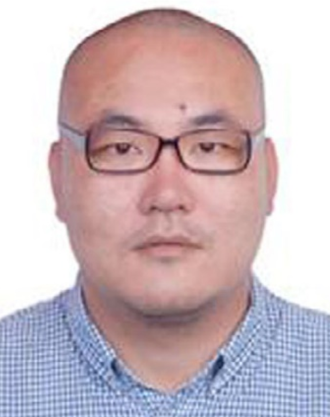}{Yueying Wang}{
(Senior Member, IEEE) received the B.S. degree in mechanical engineering and automation from Beijing Institute of Technology, Beijing, China, in 2006, and the M.S. degree in navigation, guidance, and control and the Ph.D. degree in control science and engineering from Shanghai Jiao Tong University, Shanghai, China, in 2010 and 2015, respectively. He is currently a Full Professor with the School of Mechatronic Engineering and Automation, Shanghai University, Shanghai. His research interests include intelligent perception, control, and decision-making of complex dynamic systems, and unmanned surface vehicles. He is an Associate Editor of IEEE Transactions on Neural Networks and Learning Systems and IEEE Transactions on Cybernetics.
}

\pacerbio{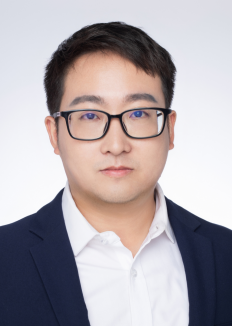}{Jie Wen}{
received the Ph.D. degree in Computer Science and Technology at Harbin Institute of Technology, Shenzhen in 2019. He is currently a Professor at the School of Computer Science and Technology, Harbin Institute of Technology, Shenzhen. His research interests include image and video enhancement, pattern recognition, and machine learning. He serves as an \textbf{Associate Editor} of \textit{IEEE Transactions on Pattern Analysis and Machine Intelligence}, \textit{IEEE Transactions on Image Processing}, \textit{IEEE Transactions on Information Forensics and Security}, \textit{IEEE Transactions on Multimedia}, \textit{IEEE Transactions on Circuits and Systems for Video Technology}, \textit{Pattern Recognition}, and an \textbf{Area Editor} of \textit{Information Fusion}. He was on the \textbf{Young Editorial Board} of \textit{CAAI Transactions on Intelligence Technology} and was an Action Editor of \textit{Transactions on Machine Learning Research}. He also served as the \textbf{Area Chair} of \textit{NeurIPS}, \textit{ICLR}, \textit{ICML}, \textit{AAAI}, and \textit{ACM MM}. For more information, please refer to the homepage: https://sites.google.com/view/jerry-wen-hit/home.
}

\clearpage
\flushbottom
\appendix
\section{Implementation Details}
\label{app:implementation}

\paragraph{Interface and visual routing.}
Following LLM-RG4, source encodings are query-compressed and arranged in a fixed frontal--lateral--previous-report layout, with unavailable lateral and previous-report branches represented by zero-valued slots before fusion \citep{wang2025llmrg4}. Images are processed with the RAD-DINO image processor at $518\times518$ resolution without data augmentation in the reported runs. The visual router uses frozen RAD-DINO features from depths $\mathcal K=\{4,8,12\}$, removes the CLS token, and preserves patch correspondence across depths. The same RAD-DINO encoder and depth router are applied to frontal and lateral views. Refine produces 1,369 non-CLS patch tokens of dimension 768 per image. The frontal branch is compressed by 128 learned visual queries through cross-attention; the resulting 128 frontal query features are reused as queries for lateral-image and previous-report compression. Previous reports are first encoded by frozen CXR-BERT \citep{boecking2022making} with a maximum input length of 100 tokens. Each compressed source is subsequently projected from 768 to the 4096-dimensional language-model space, yielding $\mathbf Z_s\in\mathbb R^{128\times4096}$ for $s\in\{f,l,p\}$.

In the reported implementation, $d_r=d_v=768$, so each alignment projection $\mathbf P_\ell\in\mathbb R^{768\times768}$ is initialized as a square identity matrix with $\mathbf b_\ell=\mathbf 0$. The affine scale and bias of each layer-specific normalization module are initialized to one and zero, respectively, while the shared routing scorer and depth biases are initialized to zero. The routed residual projection is also zero-initialized, so the initial correction is exactly zero. The residual scale is initialized to $\alpha_D=0.1$ with upper bound $\bar\alpha_D=0.5$. The routing candidates are taken from RAD-DINO hidden states at depths 4, 8, and 12, whereas the residual endpoint is the encoder's final hidden state after final normalization. The depth-12 routing candidate and the normalized endpoint preserve the same patch indexing but need not be numerically identical.

\paragraph{Prefix routing.}
The Calibrate module is inserted once at the pre-Transformer input-embedding level and is used at the same site during teacher forcing and autoregressive generation. The low-rank residual branch uses rank $r=16$ and dropout 0.05. The availability embedding and gate hidden dimension are both 4096 in the reported model. With $d=d_a=d_g=4096$, the gate projection has $\mathbf W_g\in\mathbb{R}^{4096\times12288}$.
The gate output vector $\mathbf w_g$ is initialized to zero with
$b_{\mathrm{out}}=-2.2$, while $\mathbf W_{\uparrow}$ is zero-initialized; the module therefore starts from an identity mapping with an initial sample-wise gate of approximately 0.1.

The valid-prefix mask follows the prompt attention mask and excludes padding and all teacher-forcing target-side positions. The routing gate is computed from the mean valid-prefix representation, the mean of the actually observed source representations, and the learned availability embedding. Missing lateral or previous-report sources are omitted from this observed-evidence average. A single scalar gate is produced per sample and broadcast across valid prefix positions, whereas the low-rank residual branch produces
token-specific corrections.

\paragraph{Commitment supervision.}
Clinical commitments are supervised by a hybrid offline pipeline. We construct a cache of 10,000 training records, with 2,500 records from each of SN, SW, MN, and MW. The selected records are drawn only from the training split using a fixed stratified sampling procedure that emphasizes clinically relevant patterns such as uncertainty, negation, devices, common findings, and temporal comparisons. The preserved cache contains no validation- or test-split records.

For cached records, an external model with API identifier \texttt{gpt-4.1-mini} is queried with temperature 0 and a maximum of 600 output tokens. The request contains the current reference report together with a rule-based candidate anchor and sampling metadata. The output schema contains positive, negative, and uncertain lists over the fixed 13-finding vocabulary reproduced in Appendix~\ref{app:cag-prompt}, together with a short rationale used only for validation. Outputs are normalized, restricted to the predefined vocabulary, and deduplicated across polarity fields before entering the cache.

For a training sample not covered by the cache, a deterministic rule-based extractor constructs the commitment over the same 13 findings. Reports are lower-cased, whitespace-normalized, and segmented at sentence or semicolon boundaries. For each finding, uncertainty cues such as \texttt{possible}, \texttt{probable}, \texttt{may represent}, \texttt{could represent}, \texttt{questionable}, and \texttt{suspect} are checked before local negation cues. Negation is detected within local windows around the finding using expressions such as \texttt{no}, \texttt{not}, \texttt{without}, \texttt{absent}, \texttt{negative for}, and \texttt{no evidence of}, with an exception for phrases such as \texttt{no change}. The formal fallback does not apply a separate temporal-state resolver: expressions such as \texttt{stable}, \texttt{resolved}, or \texttt{no new} are handled only through these surface matching rules. Negative matches remain candidates while subsequent mentions are inspected, whereas the first positive or uncertain match terminates the search for that finding. Support devices are matched through explicit tube, catheter, PICC, central-line, device, and pacemaker patterns. Each polarity field is deduplicated and serialized with at most eight findings; completely empty anchors use \texttt{none} for all three fields. The external model is never called online during PACER training or inference.

The commitment-first target has the form
\begin{quote}
\small\ttfamily
<ANCHOR> positive: ...; negative: ...; uncertain: ... </ANCHOR>\\
<REPORT>\\
report text\\
</REPORT></s>
\end{quote}
whereas the direct-report target contains only the cleaned report followed by \texttt{</s>}. Both routes use the same task-specific base instruction inherited from LLM-RG4. The commitment-first instruction $\boldsymbol\iota_{\mathrm C}$ appends the exact suffix \texttt{First output <ANCHOR> positive, negative and uncertain findings, then output the final report in <REPORT>.}, whereas the direct-report instruction $\boldsymbol\iota_{\mathrm D}$ appends no additional suffix. The same commitment-first instruction is used during inference. At inference, the Vicuna decoder generates the commitment and report in a single left-to-right call; neither the external model nor a cached ground-truth commitment is used as inference guidance.

\subsection{Optimization Protocol}
\label{app:training-protocol}

The two optimization stages described in the main text map to three physical training phases. Paper Stage~1 consists of an SN Refine warm-up followed by four-context Refine+Commit parent construction. Paper Stage~2 then freezes that parent and optimizes only Calibrate. Table~\ref{tab:appendix_training} summarizes the run-level settings that are directly supported by the saved checkpoints and logs.

\begin{table}[h]
\centering
\caption{Optimization protocol used for PACER. ``Batch'' denotes per-device batch size and ``Accum.'' gradient accumulation.}
\label{tab:appendix_training}
\small
\setlength{\tabcolsep}{4pt}
\begin{tabular}{@{}lccccc@{}}
\toprule
Phase & Context & LR & Batch & Accum. & Updates \\
\midrule
Refine warm-up
& SN & $3{\times}10^{-4}$ & 24 & 2 & 14,384 \\
Refine+Commit parent
& SN/SW/MN/MW & $3{\times}10^{-4}$ & 16 & 2 & 43,152 \\
Calibrate
& SN/SW/MN/MW & $1{\times}10^{-4}$ & 16 & 2 & 21,576 \\
\bottomrule
\end{tabular}
\end{table}

All three training phases were executed as single-GPU runs on NVIDIA RTX PRO 6000 Blackwell Workstation Edition GPUs. Accordingly, the effective global batch sizes, computed as per-device batch size times gradient accumulation, are 48 for Refine warm-up and 32 for both Refine+Commit and Calibrate. During Refine warm-up, the depth router and frontal visual interface are optimized while RAD-DINO, CXR-BERT, and the base Vicuna parameters remain frozen. Refine+Commit parent construction continues to train Refine and the source interfaces, introduces the commitment trajectory, and applies LoRA to Vicuna with rank 32, scaling 64, and dropout 0.1 on the effective \texttt{q\_proj}/\texttt{v\_proj} target modules. During Calibrate training, the full parent—including Refine, source interfaces, fusion modules, and Vicuna LoRA—is frozen, and only the Calibrate parameters, including the availability embeddings, gate network, and low-rank residual branch, are updated.

Trajectory-routing and commitment-loss weights are stage-specific. During Refine+Commit parent construction, the direct-report probability increases linearly from 0.1 to 0.5 over the first 16,000 optimizer updates and remains at 0.5 thereafter, with commitment-token weight
$\lambda_{\mathrm Q}=0.35$. During Calibrate training, the direct-report probability is fixed at $\rho=0.5$ and $\lambda_{\mathrm Q}=0.20$. The report-side TLW coefficient is $\lambda_{\mathrm{TLW}}=0.75$ in both stages that use the report-generation objective. Under trajectory weighting, the commitment-side span extends through the opening \texttt{<REPORT>} delimiter and its following newline, while the report-side span begins with the report content and includes the closing delimiter and EOS. The tokenizer-added BOS label is ignored by cross-entropy but retains its route-dependent weight in the normalization denominator.

\paragraph{Inherited token-level weighting.}
We reuse the report-side emphasized-token mask released with LLM-RG4 \citep{wang2025llmrg4} and incorporate it through the independently normalized auxiliary term $\mathcal L_{\mathrm{TLW}}$ defined below. In the original construction, CheXbert first identifies positive or uncertain observations, Integrated Gradients provides token attribution scores, and Gaussian smoothing is applied within the report. If an attribution score in a sentence exceeds the threshold of 0.4, the tokens of that sentence receive the elevated coefficient 1.75 rather than the default coefficient 1. In the released annotations used here, these emphasized report positions are stored as binary \texttt{new\_scores}. Let $s_{bt}\in\{0,1\}$ denote this stored mask after alignment to the report span. The auxiliary term used by PACER is
\[
\mathcal L_{\mathrm{TLW}}
=
\frac{\sum_{b,t\in\Omega_Y^{(b)}} s_{bt}\,\ell_{bt}}
{\sum_{b,t\in\Omega_Y^{(b)}} s_{bt}+\varepsilon}.
\]
PACER preserves the inherited report-side scores but shifts them to the report positions of the routed trajectory; commitment positions have $s_{bt}=0$ and are supervised separately through the trajectory loss. We use $\lambda_{\mathrm{TLW}}=0.75$ and $\varepsilon=10^{-3}$.

\paragraph{Calibrate regularization.}
During the final Calibrate stage, the residual-off frozen parent provides the teacher next-token distribution and the adapted model provides the student distribution on the same observed case and routed target. The consistency term uses $\mathrm{KL}(p_0\Vert p_\theta)$ with temperature 1 and coefficient $\lambda_{\mathrm{cons}}=0.05$. The consistency set $\Omega_{\mathrm{cons}}^{(b)}$ excludes the complete \texttt{<ANCHOR>...\string</ANCHOR>} prediction span and includes the subsequent report structural tokens, report content, and EOS when present. The prefix-drift term is the mean squared magnitude of the actual gated residual over valid prefix positions, with coefficient $\lambda_{\mathrm{drift}}=0.01$. We use $\varepsilon=10^{-3}$ in the weighted trajectory-loss normalization.

\paragraph{Trajectory routing and report extraction.}
Route selection precedes both instruction and target construction, so a direct-report sample contains neither commitment nor report delimiters, whereas a commitment-first sample contains the complete commitment--report trajectory. At evaluation, configurations with Commit use commitment-first decoding, whereas configurations without Commit use direct-report decoding. For the four-context evaluation, PACER uses deterministic beam search with three beams, \texttt{do\_sample=False}, 80--260 generated tokens, repetition penalty 2.0, and length penalty 2.0. The conventional findings-only SN evaluation uses five beams, \texttt{do\_sample=False}, 50--200 generated tokens, repetition penalty 2.0, and length penalty 2.15. Because commitment and report are produced by a single autoregressive generation call, the repetition penalty is applied over the complete generated history rather than being reset at the beginning of the report.

The decoded report is extracted from the text following the last \texttt{<REPORT>} marker and ends at the corresponding \texttt{</REPORT>} marker when present. If a valid opening marker is absent or the extracted report is empty, the raw decoded generation is retained as a fallback. We audited all reported PACER outputs under both the four-context MIMIC-RG4 evaluation and the conventional SN evaluation. Every output contained a valid opening marker and yielded a nonempty extracted report, so the raw-generation fallback was never invoked for the reported CE or NLG results.

\subsection{Preserved Commitment-Labeling Prompt}
\label{app:cag-prompt}

For transparency, we reproduce below the preserved system prompt associated with the offline commitment-labeling pipeline. The instruction content and ordering are preserved; the internal implementation header is normalized for presentation, and line wrapping is adjusted only for typesetting.

{\small
\begin{verbatim}
# Clinical Commitment Labeling Prompt

You are labeling chest X-ray report facts for a radiology report generation
experiment. Your task is to produce a compact clinical anchor from the
provided report. The anchor is used for training only. Do not add facts that
are not supported by the report.

Allowed finding vocabulary:

- atelectasis
- cardiomegaly
- consolidation
- edema
- enlarged mediastinum
- fracture
- lung lesion
- lung opacity
- pleural effusion
- pleural abnormality
- pneumonia
- pneumothorax
- support devices

Output exactly one JSON object with this schema:

{
  "sample_uid": "<copy from input>",
  "positive": ["finding_name"],
  "negative": ["finding_name"],
  "uncertain": ["finding_name"],
  "rationale": {
    "finding_name": "short evidence phrase copied or paraphrased from report"
  }
}

Rules:

1. Use only the allowed finding names. Do not invent new labels.
2. A finding must appear in at most one of `positive`, `negative`, or
   `uncertain`.
3. Put a finding in `positive` only when the report states it is present.
4. Put a finding in `negative` only when the report explicitly denies that
   specific finding or a very direct synonym of that finding.
   - Good: "no pleural effusion" -> pleural effusion negative.
   - Good: "no focal consolidation" -> consolidation negative.
   - Good: "heart size is normal" -> cardiomegaly negative.
   - Bad: "lungs are clear" -> do not automatically list every lung finding
     as negative.
   - Bad: a finding is not mentioned -> do not label it negative.
   - Bad: "no support devices mentioned" -> do not label support devices
     negative.
   - Bad: "normal chest" -> do not enumerate all findings as negative.
   - Bad: "mediastinal and hilar contours are normal" -> do not list
     unrelated lung findings as negative.
5. Put a finding in `uncertain` for possible/probable/suspected/
   difficult-to-exclude findings, including wording such as:
   - possible
   - possibly
   - may represent
   - could represent / could reflect / could be seen with
   - cannot be excluded / difficult to exclude / not excluded
   - suspicious for / concerning for / worrisome for
6. `no new`, `unchanged`, `stable`, and `no interval change` do not mean
   negative. If the finding is still described as present, label it positive.
   If the report only says no new finding but does not state a current finding
   is present, do not label that finding.
7. `resolved`, `cleared`, `resolution of`, and `no longer seen` usually mean
   the finding is currently absent. Put it in `negative` only if the wording
   clearly says it has resolved or is no longer present.
8. Support devices include tubes, catheters, PICC lines, central lines, ports,
   pacemakers, ICDs, leads, drains, chest tubes, and feeding/enteric/NG/ET
   tubes.
9. Do not mark `support devices` positive just because the report says support
   devices are unchanged if no device is named.
10. Do not mark `fracture` positive when the report says no fracture or no
    displaced fracture.
11. Hydropneumothorax should imply `pneumothorax` positive and usually
    `pleural effusion` positive unless the report clearly separates them.
12. Pleural thickening/scarring/plaques should be `pleural abnormality`, not
    `pleural effusion`, unless fluid/effusion is also stated.
13. Use `lung lesion` for nodules, masses, lesions, or metastatic pulmonary
    nodules.
14. Use `lung opacity` for opacities, infiltrates, airspace disease, or
    opacification. If the report says opacity may represent pneumonia, put
    `lung opacity` positive and `pneumonia` uncertain.
15. Keep each list concise and clinically important. Empty lists are allowed.
16. The negative list should usually be short. Do not enumerate all absent or
    unmentioned findings. For this training anchor, unknown/unmentioned means
    omitted, not negative.
    Prefer negative labels only for directly denied high-value findings such
    as pneumothorax, pleural effusion, consolidation, edema, pneumonia,
    fracture, cardiomegaly, or enlarged mediastinum.
    Do not put `lung opacity`, `lung lesion`, `atelectasis`,
    `support devices`, or `pleural abnormality` in negative unless those exact
    concepts are clearly denied in the report.
    If the negative list would exceed 6 findings, keep only the most explicit
    direct negations and omit broad inferred negatives.
17. If a report says vascular congestion, elevated pulmonary venous pressure,
    pulmonary vascular engorgement, or fluid overload, label `edema` positive
    or uncertain depending on certainty. Do not label `edema` negative.
    - If the phrase is hedged, such as "could reflect elevated pulmonary
      venous pressure", label `edema` uncertain.
    - If the report explicitly says "pulmonary edema" or "vascular
      congestion", label `edema` positive.
18. If a report says "no focal consolidation concerning for pneumonia",
    label `consolidation` negative and `pneumonia` negative, not uncertain.
19. If the report says the mediastinum is widened or the cardiomediastinal
    silhouette is enlarged, label `enlarged mediastinum` positive.
20. If the report says heart size is normal, you may label `cardiomegaly`
    negative. If the report says mediastinal contours are normal, you may
    label `enlarged mediastinum` negative. Do not use these normal statements
    to infer other negative findings.
21. Never copy a finding into multiple state lists. If evidence is mixed,
    choose the most clinically accurate single state in this priority:
    positive if definitely present; uncertain if possible/suspected;
    negative if explicitly denied and not also described as present.

Input format:

{
  "sample_uid": "...",
  "scenario": "sn|sw|mn|mw",
  "report": "...",
  "regex_anchor": {
    "positive": [],
    "negative": [],
    "uncertain": []
  },
  "risk_buckets": []
}

Return only the JSON object. No markdown and no explanation outside JSON.
\end{verbatim}
}

\section{Evaluation Protocol and Result Provenance}
\label{app:evaluation-notes}

\subsection{Dataset Splits and Availability Settings}
\label{app:data-splits}

We use the existing MIMIC-RG4 train/validation/test splits rather than creating a new random split. Table~\ref{tab:appendix_splits} reports the scenario-record counts used by the four availability settings. Records can correspond to the same underlying examination across multiple availability states, so the sum across scenarios should not be interpreted as the number of independent examinations. An audit of the current annotations found no direct train--validation, train--test, or validation--test overlap at the study or subject level.

\begin{table}[h]
\centering
\caption{Scenario-record counts in the four MIMIC-RG4 availability states.}
\label{tab:appendix_splits}
\small
\begin{tabular}{@{}lrrr@{}}
\toprule
Context & Train & Validation & Test \\
\midrule
SN & 172,608 & 1,391 & 2,357 \\
SW & 112,776 & 937 & 2,026 \\
MN & 91,341 & 701 & 1,004 \\
MW & 47,686 & 371 & 828 \\
\bottomrule
\end{tabular}
\end{table}

The structured availability axes are only the lateral radiograph and previous report. Indication/history may be included in the prompt when available but is not treated as an availability axis. The previous-report source in SW/MW is encoded with frozen CXR-BERT; a previous image is not used as an additional model input in the PACER configuration reported here.

\subsection{Clinical and Language Metrics}
\label{app:metrics}

Clinical efficacy is computed with the repository CheXbert evaluator over 14 report labels. Blank/negative outputs are treated as absent, while uncertain and positive outputs are treated as present for the reported CE scores. Precision, recall, and F1 are micro-averaged over cases and labels. Four-context ablation summaries are arithmetic means of the four per-context metrics, giving equal weight to SN, SW, MN, and MW rather than re-micro-averaging all scenario records jointly.

Language metrics use the repository evaluation implementation for BLEU-1--4, ROUGE-L, and METEOR 1.5. The reported four-context results are scored against the complete cleaned MIMIC-RG4 references used by the formal evaluation pipeline.

\subsection{Four-Context Result Provenance}
\label{app:four-context-provenance}

Table~\ref{tab:main_rg4} reports the complete CE and NLG comparison under the four MIMIC-RG4 input contexts. The published CXRMate, RadFM, and LLM-RG4 values follow the corresponding four-context comparison in \citet{wang2025llmrg4}. SimMLM and RAGPT are our local adaptations described
in Appendix~\ref{app:baseline-adaptation}, and the PACER row is obtained from our formal four-context evaluation.

Per-context values are reported to three decimal places. Ablation averages are computed from unrounded per-context outputs and may therefore differ by 0.001 from values recomputed from the rounded entries in Table~\ref{tab:main_rg4}. The \emph{Base} row in Table~\ref{tab:core_ablation} is the controlled local ablation parent, whereas Table~\ref{tab:main_rg4} reports the published LLM-RG4 result.

\subsection{Conventional SN Evaluation}
\label{app:conventional-sn}

For this fixed-availability findings-only setting, PACER is trained separately with the Refine+Commit configuration, without Calibrate. Table~\ref{tab:sn_findings} separates literature-reported results under their original protocols from results evaluated on the common findings-only SN test set. Clean NLG metrics use the cleaned findings references, whereas Original NLG metrics use the corresponding original MIMIC-CXR findings reports. Because the upper block preserves the original evaluation protocols, it is included for broader literature context rather than direct ranking; direct comparisons in Section~\ref{sec:sn} refer to the common-test block. CE values in the literature-only block retain the metric definitions and aggregation protocols reported by the original studies; our unified micro-averaged CheXbert evaluation applies to the common-test evaluation.

The common-test baselines through LLM-RG4 follow Table~2 of \citet{wang2025llmrg4}. The literature rows through InVERGe use the same comparison set, with KiUT BLEU-1 and EKAGen BLEU-4 reported according to their original sources. REVTAF, ESC-RL, and S2D-Align are additional recent literature references \citep{zhou2025revtaf,zhou2026escrl,gao2026s2dalign}. The additional MambaXray-VL-Large and CheXOne results are produced locally under the protocols detailed in Appendix~\ref{app:recent-sn-baselines}.

\section{Local Adaptations and Ablation Evaluation}
\label{app:baseline-adaptation}

\subsection{Incomplete-Multimodal Baseline Adaptations}
\label{app:incomplete-baselines}

The four-context SimMLM and RAGPT rows in Table~\ref{tab:main_rg4} are local mechanism-transfer adaptations of \citet{li2025simmlm} and \citet{lang2025ragpt}, respectively, to the MIMIC-RG4 structured incomplete-context setting. They are not values reported in the original papers and should not be interpreted as full official end-to-end reproductions.

For SimMLM, we retain a frozen common RRG parent and introduce source-specific residual experts for frontal, lateral, and previous-report representations. A learned availability-aware gate mixes the observed experts, and training uses paired more-evidence/fewer-evidence examples with a ranking term while retaining the parent report-generation objective. For RAGPT, we transfer the published missing-modality generation and contextual adaptive prompting modules to the frontal/lateral/previous-report source layout; the adapted model uses a train-only retrieval bank and trains only the added RAGPT modules while the common parent remains frozen.

Both adaptations use the same canonical four-context test populations and the same CE/NLG evaluator used for the PACER comparison. Their final adaptation budgets are fixed at 20,000 optimizer updates with seed 42. 

\paragraph{Additional checkpoint-transfer and common-test evaluations.}
We additionally evaluated several released checkpoints and implementations as supplementary transfer references using the same frozen CE/NLG evaluator. The released MLRG checkpoint \citep{liu2025mlrg} is evaluated without local parameter updates. Because it was trained for MLRG's original multi-view longitudinal setting rather than the four-context MIMIC-RG4 training protocol, we do not include it in the direct ranking of Table~\ref{tab:main_rg4}. In addition to its SN result reported in Table~\ref{tab:additional_transfer}, its arithmetic mean over the four MIMIC-RG4 contexts is 0.535/0.394/0.454 for P/R/F1 and 0.297/0.080/0.250 for B@1/B@4/R-L.

We further evaluate RGRG \citep{tanida2023region}, EKAGen \citep{bu2024ekagen}, and CheXmix \citep{kumar2026chexmix} on the common findings-only SN test population. For RGRG, the official beam-search configuration is treated as the primary transfer setting, while greedy decoding and complete-deduplication are included only as inference-sensitivity variants. For EKAGen, 11 test cases have unavailable knowledge-base entries; we therefore report both mapping these cases to empty outputs and a nearest-neighbor handling variant. CheXmix produces empty native Findings for 1,217 cases, contributing to its substantially lower transfer performance. The complete supplementary SN results are summarized in Table~\ref{tab:additional_transfer}.
\begin{table}[t]
\centering
\caption{
Supplementary checkpoint-transfer and common-test SN evaluations.
RGRG decoding settings and EKAGen missing-entry handling are protocol variants rather than independent models.
}
\label{tab:additional_transfer}
\small
\setlength{\tabcolsep}{3.2pt}
\begin{tabular}{@{}lcccccc@{}}
\toprule
Method & P & R & F1 & B@1 & B@4 & R-L \\
\midrule
MLRG checkpoint
    & 0.536 & 0.402 & 0.460 & 0.365 & 0.097 & 0.269 \\
\midrule
RGRG (official beam4)
    & 0.515 & 0.523 & 0.519 & 0.366 & 0.105 & 0.264 \\
RGRG (greedy)
    & 0.489 & 0.603 & 0.540 & 0.362 & 0.109 & 0.271 \\
RGRG (dedup.)
    & 0.517 & 0.515 & 0.516 & 0.294 & 0.093 & 0.254 \\
\midrule
EKAGen (missing$\rightarrow$empty)
    & 0.528 & 0.442 & 0.481 & 0.346 & 0.089 & 0.275 \\
EKAGen (nearest-neighbor)
    & 0.529 & 0.444 & 0.483 & 0.346 & 0.089 & 0.276 \\
\midrule
CheXmix
    & 0.180 & 0.077 & 0.108 & 0.045 & 0.014 & 0.142 \\
\bottomrule
\end{tabular}
\end{table}

\subsection{Additional Conventional-SN Baselines}
\label{app:recent-sn-baselines}

\paragraph{MambaXray-VL-Large.}
The reported MambaXray-VL-Large result \citep{wang2025cxpmrg} is obtained by locally training its Stage-3 downstream model from the released \texttt{MambaXrayCLIP-L.pth} Stage-2 visual weights with Llama-2-7B-chat. The language model is frozen while the visual encoder, projection, and normalization modules are optimized using AdamW with learning rate $10^{-4}$, weight decay 0.01, batch size 6, no gradient accumulation, and seed 42. Training runs for five epochs (111,545 optimizer updates) with cosine annealing to a minimum learning rate of $10^{-6}$. Checkpoint selection maximizes $0.8\,\mathrm{BLEU4}+0.2\,\mathrm{CIDEr}$ on validation data. Evaluation uses deterministic three-beam decoding with 80--120 generated tokens and repetition/length penalties of 2.0/2.0.

\paragraph{CheXOne.}
The CheXOne result uses the released \texttt{StanfordAIMI/CheXOne} checkpoint \citep{zhang2026chexone} with no local parameter updates. The inference prompt is \texttt{Write an example findings section for the CXR. Please reason step by step, and put your final answer within \detokenize{\boxed{{}}}.} Generation is deterministic with one beam and at most 1,024 new tokens. The final findings are extracted from the last \texttt{\detokenize{\boxed{...}}} block using brace-depth matching rather than a regular expression; an absent or unclosed box yields an empty extraction. Both baselines are scored with the same frozen SN evaluator used for the common-test block of Table~\ref{tab:sn_findings}.

\subsection{Ablation Evaluation and Shared Configurations}
\label{app:ablation-protocol}

Component and mechanism ablations are summarized by the arithmetic mean over the four MIMIC-RG4 contexts. Table~\ref{tab:core_ablation} starts from a local ablation parent and evaluates Refine alone, Commit alone, their combination, and the final Calibrate stage. Table~\ref{tab:mechanism_ablation} varies one mechanism group at a time. Refine variants use the same Commit training protocol with Calibrate disabled; Commit variants use the same Refine architecture and the checkpoint obtained after the SN Refine warm-up with Calibrate disabled; and Calibrate variants are trained from the same frozen Refine+Commit parent.

Shared control configurations are intentionally repeated across mechanism groups for within-group comparison. Consequently, identical values for \emph{Patchwise depth routing}, \emph{Stochastic trajectory routing}, and \emph{No Calibrate (no prefix routing)} correspond to the same evaluated Refine+Commit control configuration rather than independent configurations that happened to produce the same rounded scores. Global depth weighting learns a single set of softmax-normalized depth weights shared across all samples and spatial positions. For the Calibrate input ablations, all gate variants use the same architecture and parameter count; disabled cues are replaced by zero vectors while the full 12,288-dimensional gate input is retained.

\section{Discussion and Limitations}
\label{app:discussion}

The results reinforce a key distinction in flexible-context RRG: supporting heterogeneous input combinations does not by itself ensure effective evidence utilization. The consistent gains across availability states suggest that adaptation should extend beyond input fusion to visual representation, language-model conditioning, and report generation.

Several limitations remain. Our evaluation is centered on MIMIC-RG4 and the conventional MIMIC-CXR setting, leaving generalization to other institutions and less structured forms of incomplete or corrupted context for future study. Clinical efficacy follows the benchmark CheXbert binarization and therefore does not directly measure the full three-way polarity distinction among positive, negative, and uncertain findings, nor finer-grained properties such as anatomical grounding or temporal relations. The commitment ablations instead evaluate the downstream effect of polarity-structured supervision on report generation, while Figure~\ref{fig:case_study} provides qualitative evidence of temporal evidence utilization. Finally, commitment supervision is constructed with a hybrid offline pipeline that combines GPT-assisted labels for a stratified 10k subset with deterministic rule-based fallback targets for the remaining training records. Future work should evaluate expert-verified commitment supervision, polarity-sensitive and relation-aware clinical metrics, and external clinical distributions.

\end{document}